# Reasoning about Action:
# An Argumentation-Theoretic Approach

**Quoc Bao Vo**                                            VQBAO@CS.RMIT.EDU.AU
*School of Computer Science and Information Technology*
*RMIT University*
*GPO Box 2476V, Melbourne, VIC 3001, Australia*

**Norman Y. Foo**                                          NORMAN@CSE.UNSW.EDU.AU
*Knowledge Systems Group*
*Artificial Intelligence Laboratory*
*School of Computer Science and Engineering*
*University of New South Wales, Sydney, NSW 2052, Australia*

## Abstract

We present a uniform non-monotonic solution to the problems of reasoning about action on the basis of an argumentation-theoretic approach. Our theory is provably correct relative to a sensible minimisation policy introduced on top of a temporal propositional logic. Sophisticated problem domains can be formalised in our framework. As much attention of researchers in the field has been paid to the traditional and basic problems in reasoning about actions such as the frame, the qualification and the ramification problems, approaches to these problems within our formalisation lie at heart of the expositions presented in this paper.

## 1. Motivation and Introduction

The need for a good reasoning about action formalism is apparent for research in artificial intelligence (AI). Alongside the logicist point of view to artificial intelligence, more recently, there emerges the cognitivist and situated action-based approaches(see Kushmerick, 1996 and the references therein). The latter approaches provide some immediate and practical answers to certain issues of AI. The current problem domains for (Soccer) Robot Cup seem to be an area where these approaches promise to gain fruitful results. On the other hand, the logicist approach aims at long term solutions for the general problems of AI. From a logicist approach, formalising dynamic domains for reasoning about action can be realised within a logical knowledge representation. The general idea is that intelligent agents should be able to represent all kinds of knowledge in a uniform way such that some general problem solver can fully employ and find a solution based on their knowledge. As it turns out, there are difficulties with such a general approach to AI. Consider the task of formalising dynamic domains in some logical language. To formalise the dynamics of an action (or event) in a language with $n$ fluents[1], one will need to axiomatise not only about the fluents that are effected by the action but also about those that are not. Essentially, it requires that $n$ axioms be asserted. Such a formalisation can hardly be considered a good representation. Hence, there is the need to solve this problem in logic-based reasoning about action formalisms. This is the well known *frame problem* as introduced by McCarthy and Hayes (1969). Moreover, there is still a problem in axiomatising the effects of an action, called the *effect axioms*. A logical axiomatisation requires that the conditions under which the effects will take place after executing the action be precisely specified. However, there are potentially infinitely many such conditions, some of which the reasoner may never have thought about. No realistic formalisation would ever be able to exhaustively enumerate all of those conditions. Nonetheless, to start a car, most people only worry about whether they have the key to that car. They never bother checking whether there is something blocking the tailpipe or checking all electric circuits to make sure that they are all well connected. Such a story has long been well-known within the community of commonsense reasoning, in particular reasoning about action. This is known as the *qualification problem* and was introduced by McCarthy (1977).

---

1. *fluent* is a technical term referring to functions or predicates whose values can be varied relative to time.

While there have been a number of solutions to the frame problem (e.g., Shanahan, 1997; Reiter, 1991; Castilho, Gasquet, & Herzig, 1999), the qualification problem has largely been ignored with the notable exception of Thielscher's (2001) solution within the Fluent Calculus and Doherty and Kvarnström's (1998) circumscription-based solution using fluent dependency constraints. Some people argue that the frame problem is already very challenging and it would be a good approach to thoroughly solve the frame problem before complicating a formalism with the qualification problem. We argue that there is a danger of approaching these problems from that point of view for (at least) two reasons:

1. *It may be very hard to come up with a uniform solution for all problems:* while many existing solutions for the frame problem are monotonic (e.g., Reiter, 1991; Castilho et al., 1999), the qualification problem inherently requires a non-monotonic solution. This is the case with the original qualification problem as stated by McCarthy (1977) for the dynamics of actions/events need to be finitely axiomatisable and when an unexpected qualification for an action arises, the agent must necessarily retract his initial expectation that the effects caused by the action would take place, making the underlying reasoning machinery non-monotonic (see section 1.2 for a discussion on the qualification problem).

2. *Many solutions to the frame problem can only succeed under some precise assumptions.* For instance,

   - Actions always succeed. This is the *action omniscience* assumption. More precisely, this assumption dictates that the qualification problem is skipped. This is the case with all monotonic solutions to the frame problem.[2]

   - Fluents change if and only if the reasoner knows that there exists an action that possibly changes its value. This can be termed as *domain omniscience* assumption. It assumes that the reasoner has complete (ontological) knowledge of the domain about which he is reasoning.

---

2. The argument that any solution to the frame problem which works with nondeterministic action is not subject to this assumption does not stand. The quick fix of allowing such an approach to a fortiori express actions that may fail by representing failure as a possible effect is an invalid one. It is because we can no longer infer that, in the absence of evidences that suggest otherwise, actions would normally succeed. It's also worth noting that Lin's (1996) extension of Reiter's (1991) solution to the frame problem in the Situation Calculus to deal with nondeterministic action is based on circumscription.

The above two reasons are of course closely related as the former arises due to the underlying assumptions in the latter which no longer holds once the qualification problem is taken into consideration.

In the remainder of this section, we review several works on this topic before introducing the reader to our approach.

## 1.1 The Frame Problem

In the late 1960s, the frame problem had been recognised as a major obstacle to formalising dynamic domains (see the discussions and exposition by McCarthy & Hayes, 1969; Green, 1969). Several alternative responses to the frame problem have been proposed along way. To respond to the explosive number of axioms required for theorem proving-based planners as proposed by Green (1969), Fikes and Nilsson (1971) introduce procedures that operate on special data structures used to represent dynamic domains. However, for complex and sophisticated problem domains, e.g. those with domain constraints, concurrent actions, observations at different time points, etc. STRIPS quite often fails to express the domain knowledge. In fact, the expressivity of STRIPS is quite limited as has been pointed out by Lifschitz (1987). Another response attributing the frame problem as an artefact of the situation calculus has proved to be ungrounded. There are attempts to distinguish a logical or epistemological aspect of the frame problem from the computational aspect (e.g., McDermott, 1987; Kowalski, 1992). While the computational inefficiency associated with a representation of dynamic domains in the situation calculus can be attributed to the explosive number of global situations required by the situation calculus as argued by Kowalski and Sergot (1986), the logical aspect of the frame problem is inherent to any logic-based representation of dynamic domains. It is thus essential that a logical approach to AI and knowledge representation have a decent solution to the frame problem.

Later, with the introduction of the qualification problem by McCarthy (1977), it is reckoned that formalising dynamic domains not only is about solving the frame problem but would require systematic studies to fundamental issues of knowledge representation. In the early 1980s, the frame and the qualification problem were considered to be instances of commonsense reasoning problems. In particular, many believed that a non-monotonic reasoning framework would solve the frame problem. It was argued that the 'principle of inertia'[3] which is considered to be the key to the frame problem can be formalised in terms of default rules or default axioms in default reasoning. Moreover, it was also argued that to solve the qualification problem, the following common sense law should be rendered: *"an action, by default, would qualify to succeed and bring about the intended effects unless there is known reason for it not to,"* in formalisations of dynamic domains.

With the introduction of several non-monotonic reasoning formalisms, e.g. truth maintenance systems (TMSs) by Doyle (1979), default logics by Reiter (1980), circumscriptive approaches by McCarthy (1980), modal non-monotonic logic by McDermott and Doyle

---

3. The *principle of inertia* or the *common sense law of inertia* basically states that *"By default a fluent is assumed to persist over time unless there is evidence to believe otherwise."* The reader is referred to Shanahan's (1997) book for more details on this principle and the issues around it.

(1980), autoepistemic logic by Moore (1985), etc., it was believed that these problems were solved. The solutions to these problems were illustrated as examples for the proposed non-monotonic reasoning frameworks. For instance, McCarthy (1986) showed how circumscription is used to solve the frame problem relative to the *blocks world* domain. Unfortunately, Hanks and McDermott (1987) show that these formalisations do not work correctly in a simple dynamic domain known as the Yale Shooting Problem (YSP). We will now review successful attempts to solve the frame problem:

1. Baker (1989) successfully modifies the original (and incorrect) circumscription policy proposed by McCarthy (1986) to deal with the Yale Shooting Problem. In the traditional circumscriptive policy, the predicate *Abnormal* is minimised with the predicate *Holds* allowed to vary. Baker suggests that, instead of allowing *Holds* to vary, the function *Result* should be the one to be varied. While this does not solve the frame problem in its full generality, this initiates the line of research which brings many fruitful results to reasoning about action community. For a more detailed discussion about these solutions and the follow up works, the reader is referred to Shanahan's (1997) book. Furthermore, Foo, Zhang, Vo, and Peppas (2001) present an exposition on the issue from an automata and system theory point of view. In order for Baker's (1989) solution to work correctly, additional axioms need to be introduced, e.g. domain closure axioms, axioms about the existence of situations, etc. This emphasises that: (i) the circumscriptive approach to reasoning about action only works under careful designation and considerations of the domain; and more importantly, (ii) circumscription is domain dependent. That is, a domain dependent circumscriptive policy is required to correctly render the common sense of a particular problem domain.

2. A number of researchers argue that in many cases, a monotonic solution to the frame problem will be sufficient. Pednault (1989) assumes that the effect of actions on fluents are specified by *effect axioms* of the following forms:

$$\varepsilon_F^+(\vec{x}, \vec{y}, s) \supset F(\vec{x}, do(A(\vec{y}), s)), \tag{1}$$
$$\varepsilon_F^-(\vec{x}, \vec{y}, s) \supset \neg F(\vec{x}, do(A(\vec{y}), s)), \tag{2}$$

Here, $A(\vec{y})$ and $F(\vec{x}, s)$ are the parameterised action and fluent, respectively; $\varepsilon_F^+(\vec{x}, \vec{y}, s)$ and $\varepsilon_F^-(\vec{x}, \vec{y}, s)$ are first order formulas whose free variables are among $\vec{x}, \vec{y}, s$. Pednault (1989) makes the following **Causal Completeness Assumption**:

> The axioms (1) and (2) specify all the causal laws relating the action $A$ and the fluent $F$.

Note that the Causal Completeness Assumption is a stronger form of the domain omniscience assumption presented above. Under this assumption, the following frame axioms can be introduced:

$$F(\vec{x}, s) \wedge \neg \varepsilon_F^-(\vec{x}, \vec{y}, s) \supset F(\vec{x}, do(A(\vec{y}), s)).$$

and

$$\neg F(\vec{x}, s) \wedge \neg \varepsilon_F^+(\vec{x}, \vec{y}, s) \supset \neg F(\vec{x}, do(A(\vec{y}), s)).$$

Schubert (1990), elaborating on a proposal of Haas (1987), employs the so-called **Explanation Closure Axioms** of the following forms:

$$F(\vec{x}, s) \wedge \neg F(\vec{x}, do(A, s)) \supset \alpha_F(\vec{x}, A, s), \qquad (3)$$
$$\neg F(\vec{x}, s) \wedge F(\vec{x}, do(A, s)) \supset \beta_F(\vec{x}, A, s), \qquad (4)$$

Or, equivalently, we can rewrite the above two axioms as follows:

$$F(\vec{x}, s) \wedge \neg \alpha_F(\vec{x}, A, s) \supset F(\vec{x}, do(A, s)).$$

and

$$\neg F(\vec{x}, s) \wedge \neg \beta_F(\vec{x}, A, s) \supset \neg F(\vec{x}, do(A, s)).$$

Schubert's proposal is correct under the following assumption, called the **Explanation Closure Assumption**:

$\alpha_F$ *completely characterises all those actions $A$ that can cause the fluent $F$'s truth value to change from* **true** *to* **false***; similarly for $\beta_F$.*

Reiter (1991) then combines the merits of the above two proposals by systematically generating the frame axioms as proposed by Pednault (1989) with the quantifiers over the set of actions as proposed by Haas (1987) and Schubert (1990).

Other researchers who also propose monotonic solution to the frame problem include Castilho et al. (1999), and Zhang and Foo (2002).

3. Attempts to solve the frame problem using default logic (Reiter, 1980) also encounter some problematic issues. Hanks and McDermott (1987) show that a natural formulation of the Yale Shooting Problem in default logic suffers the same problem as that with circumscriptive approaches, *viz.* the existence of anomalous extensions. Morris (1988) proposes a slight modification on Hanks and McDermott's original formulation to the Yale Shooting Problem in an attempt to avoid the anomalous extensions. As pointed out by Turner (1997), Morris' formulation is complete, and thus eliminates the anomalous extensions in the Yale Shooting Problem, but unsound. More importantly, from Morris' formulation it is not clear how dynamic domains should be formulated in general. Turner (1997) himself then proposes a way to formalise dynamic domains using default logic (and also logic programming). His solution is based on the following observation: In the Yale Shooting domain and similar dynamic domains, the anomalous extensions arise because undesired effects of an action can be derived by reasoning "backward in time." For instance, in the Yale Shooting domain, by making the counterintuitive supposition that the victim of the shooting is somehow still alive after the shooting, an anomalous extension come up in the following way. First, it allows the default saying that the victim persists to be alive regarding the shooting action to be applicable. As a consequence, the gun must not be loaded before the shooting action. Therefore, it blocks the application of the default saying that the gun persists to be loaded regarding the waiting action. In other words, the loaded gun would get unloaded (magically) during the waiting action which is an undesirable conclusion. To block these lines of "backward" reasoning, Turner appeals to the non-contrapositivity of inference rules and replaces the implications by inference rules. To guarantee that this formulation work correctly, additional techniques are required such as a fact $\varphi$ will be formulated as an inference rule:

$$\frac{\neg \varphi}{\textbf{false}}$$

and enforcing the completeness of the initial situation by adding the following rules:

$$\frac{: Holds(f, S_0)}{Holds(f, S_0)} \qquad \frac{: \neg Holds(f, S_0)}{\neg Holds(f, S_0)}$$

for every fluent $f$.

However, Turner's (1997) formulation is still fairly *ad hoc* as different techniques are added to fix the known issues. For example, the inference rules are used in place of implications to block the application of the undesirable "backward" reasoning, the rules completing the initial situations are added to overcome the unsoundness issue in Morris' (1988) formulation, etc. This also shows the problematic side of default logic as a uniform formalisation to various problems of common sense reasoning. This becomes a serious issue if one proceeds with the question of how the qualification problem, a typical problem of default reasoning, is solved in Turner's (1997) formalisation of dynamic domains. This is the case, for example, when instead of asserting that the victim dies whenever it is shot by a loaded gun the reasoner can only maintain that as a default proposition as there may be many hidden possible conditions under which the victim may not die. Thus the reasoner is able to deal with 'surprising' situations in which the victim is observed to be still alive after the shooting action (of a loaded gun). Another, symmetric, case is with 'surprises' regarding the persistence of fluents. For instance, after a waiting action, which is not supposed to unload a gun, the gun, which was loaded before the wait action, is observed to be unloaded. Such a scenario was first introduced by Kautz (1986) in a scenario called the Stolen Car Problem. In both cases, reasoning "backward in time" is necessary. It is not clear how these will be rendered in Turner's formulation which explicitly intends to block "backward" reasoning. These scenarios will be analysed in the solution we present later in this paper.

## 1.2 The Qualification Problem

While there have been several solutions to the qualification problem, none of these addressed the original qualification problem introduced by McCarthy (1977) and later formalised by Ginsberg and Smith (1988).

1. Lin and Reiter (1994) propose a formalisation for action theories in the situation calculus (SC). Their formalism is an extension of Reiter's (1991) solution to the frame problem (sometimes) by incorporating state constraints. They discover that there are at least two different kinds of state constraints which they call ramification and qualification constraints. They then go further to claim a solution to the qualification (and the ramification) problem. The basic idea behind their solution to the qualification problem is that certain state constraints imply implicit preconditions of some actions. Thus an action may not be qualified even though it appears (from the explicit action description) to be. While this is of course a special case of the qualification problem, the classical qualification problem as introduced by McCarthy (1977) has a much broader extent. In this setting, the qualification problem is more a pragmatic issue than a technical issue. Similar to the frame problem, it is impractical, and sometimes impossible, to axiomatise all the possible preconditions of an action. For example, in addition to the requirement that the gun be loaded, to guarantee that performing the action shoot would kill the victim, many preconditions must also be included such as: the gun is not malfunctioning, the shooter does not miss the victim, the victim does not wear a bullet-proof jacket, etc. among which some may be very improbable such as: "no alien interferes with the bullet." The reasoner simply does not want to consider these conditions by assuming that they are not the case unless there are explicit evidences stating otherwise. In other words, the qualification problem in its original form requires that the reasoner be able to tolerate the mistaken conclusions possibly jumped to by previous inferences and to correct them appropriately. Henceforth, we will always refer to the *qualification problem* in this original form. It is this similarity to the frame problem that led John McCarthy to conjecture that:

> The *frame problem* may be a sub-case of what we call the *qualification problem*, and a good solution of the qualification problem may solve the frame problem also.
>
> (McCarthy, 1977, p. 1040, *italics are original.*)

Roughly 10 years after his introduction of the qualification problem, McCarthy (1986) presented his solution to the problem using his non-monotonic formalism of circumscription. However, this solution suffers an almost identical flaw as its counterpart regarding the frame problem: simple minimisation of abnormalities sanctions anomalous models (e.g., Thielscher, 2001).

2. McCain and Turner (1995) propose a solution to the problem described by Lin and Reiter (1994), viz. the problem of deriving the implicit preconditions from state constraints. McCain and Turner's solution is posed in a model-based representation of action theories.

3. Similar to McCain and Turner's (1995) result, Baral (1995) offers a solution to the problem defined by Lin and Reiter using a state-based representation. Baral extends the language of disjunctive logic programs for state specification and as an action description language.

4. Doherty and Kvarnström (1998) make a careful investigation to the qualification problem. They are aware of the shortcomings present in the definition of the "qualification problem" introduced by Lin and Reiter (1994). They proceed one step further to distinguish between the *weak* and *strong* forms of the qualification problem. To deal comprehensively with the qualification problem in its full extent, Doherty and Kvarnström apply circumscription on a predicate which plays a similar role to the predicate *Poss* used by Lin and Reiter (1994).

Even though Doherty and Kvarnström's (1998) solution is closest in spirit to the original form of the qualification problem, there is still a serious problem with their approach. The intended designation on predicate *Poss* and its variants is action-oriented. That is, it would qualify on the *executability* condition for the action under consideration, not towards the effects that action is supposed to cause. In other words, only circumscribing *Poss* does not guarantee to capture the full extent of the qualification problem. For example, a single action of shooting a gun may cause several effects: killing the victim, making a loud noise, emptying the cartridge, etc. The conditions for such an action to be executable is: having the gun, the gun is not broken, the gun is loadable, etc. Once the action is executable, it is not necessary that all effects will take place. It may be the case that there is a loud noise and the cartridge is emptied but the victim is still alive since the victim was wearing a bullet-proof jacket. Assuming that the reasoner is somehow aware about this possibility, should he include the requirement that the victim not wear a bullet-proof jacket as a qualification for the action shoot? Perhaps he should not for he would still expect to hear a loud noise and the cartridge to be emptied after the shoot action.

These are not the end of all the troubles though. The presence of both qualification and ramification constraints causes several complications. Firstly, they are not syntactically distinguishable. Secondly, as has been mentioned by Doherty and Kvarnström (1998), qualification constraints may cause indirect effects to arise and vice versa, i.e. ramification constraints may reveal implicit action preconditions.

REMARK: The terms *ramification constraints* and *qualification constraints* were first introduced by Lin and Reiter (1994) when a careful examination to state constraints was taken. As discussed by Doherty and Kvarnström (1998), these two kinds of constraints might interact in several ways. Consider the example introduced by Doherty and Kvarnström (1998): the only preconditions of the action *board* a plane are *having-ticket* and *at-gate*. However, to a passenger who places a gun into his pocket at home before travelling to the airport and proceeding to the gate, a new qualification for the action *board* materialises. Because one ramification of putting an object in your pocket is that it will stay with you as you travel from location to location (i.e. result of a *ramification constraint*), a reasoner could easily conjecture that our passenger fails to board the plane. On the other hand, the fact that a passenger who possesses a gun when trying to board the plane must fail to board the plane is a result of a *qualification constraint*. Now, not only the fact that a passenger possesses a gun disqualifies the action of boarding a plane but it also brings about an indirect effect when the action of boarding a plane is executed: the passenger being put under arrest. Note that for this indirect effect to take place, both requirements must be present: the action of boarding a plane being executed, and the above qualification constraint is present. In other words, a qualification constraint might also bring about indirect effects.

We believe that all these problems are too sophisticated for any circumscription policy to successfully address in most situations. Furthermore, such a policy would be extremely hard to understand and error-prone. Recall the failure of non-monotonic reasoning formalisms regarding the frame problem (in its simplest form viz. without the ramification and the qualification problems). Researchers had failed to point out this bug for several years before Hanks and McDermott discovered it in their award winning paper (1987).

5. More recently, Thielscher (2001) gives an exposition on the qualification problem. Thielscher discusses the problem sustained by McCarthy's (1986) simplistic circumscription policy, *viz.* the anomalous models. He then introduces a default logic based formalisation for the qualification problem in the Fluent Calculus and shows that his formalisation does not suffer the problem of anomalous models. Note that in Thielscher's formalisation, circumscription is still required to generate the initial theories of the default theories (in addition to the set of default rules). Nevertheless, Thielscher's solution still suffers the following drawbacks: (i) Thielscher's use of the predicate *Poss* is in the same way as has been formulated by Doherty and Kvarnström (1998). Thus, qualifications are taken over the executability conditions for the actions rather than over different effects of the actions; and (ii) while Thielscher shows that the problem of anomalous models sustained by McCarthy's (1986) circumscription policy is overcome in his formalisation, it's not entirely clear whether Thielscher's formalisation which is based on both circumscription and default logic will not suffer from other anomalies.

## 1.3 The Ramification Problem

In the context of reasoning about action, the ramification problem is mainly related to indirect effects. Finding a solution to this problem may not be easy as indirect effects indicate exceptions to frame assumptions and require special treatment. While there have been several formalisms dealing with the ramification problem, e.g., see (Lin, 1995; McCain & Turner, 1995; Thielscher, 1997), there are still several issues that need a more careful consideration. We consider three examples to motivate our discussion.

**Example 1** Consider Thielscher's (1997) circuit:

This example is interesting because it gives a counterexample for the minimalistic approaches e.g. in the work of McCain and Turner (1995). In this domain, the intended relationship between *relay* and $sw_2$ is that when *relay* is on, it would make $sw_2$ jump off. Thus, when $sw_1$ and $sw_3$ are both closed, $sw_2$ can not be also closed as that is prevented by *relay*. However, there is certainly a duration (no matter how short it is) before $sw_2$ is forced to jump off by *relay*. In the state given in Figure 1, after performing the action of closing switch $sw_1$, two next states are equally possible: one in which *detect* is on, in another it is off.[4] Only the latter is sanctioned in a minimalistic account. Through this example, Thielscher pointed out the need for keeping track of the chains of applications of indirect effects.

---

4. The reason for nondeterminism in this case is due to insufficiency of domain information: depending on the sensitivity of *relay*, *light* and *detect*, when *light* could get lit quickly and *detect* is very sensitive to

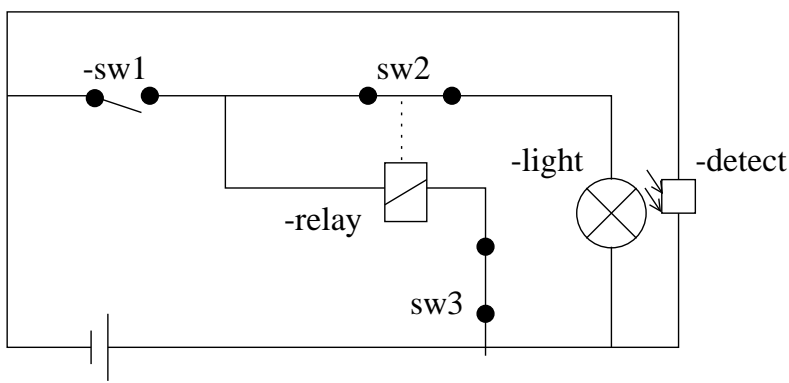

Figure 1: Thielscher's circuit

Thielscher (1997) proposes a way to remedy this problem by keeping track of the applications of the domain constraints which are re-expressed in terms of causal relationships. Thus, given the above example, his formalism is able to arrive at a next state in which the *detect* is on. Following such chains of causal relationships, the dynamic system undergoes several intermediate states before arriving at the next state.  □

In this paper, we proceed one step further from Thielscher's (1997) position by formally representing the intermediate states as possible states of the world.[5] We believe that an intelligent agent should be able to reason about these intermediate states even though they may not satisfy all domain constraints. This capability is especially important if the reasoner needs to explain certain observations about the world in a systematic way. We note that given an observation, there may be several chains of causal relationships that bring about that observation.[6] Unless intermediate states are explicitly represented and reasoned about, there is no way for an agent to have a full insight to the system in hand and certain information would be missing.

**Example 2** Consider Lin's (1995) spring-loaded suitcase with two latches. Let's assume that the latches can be toggled if the suitcase is closed. The following state constraint is supposed to apply in this domain: $up(Latch_1) \wedge up(Latch_2) \supset open(Suitcase)$.

The question is: *how does a robot close the above suitcase back after opening it?* McCain and Turner (1997) also consider this problem and their answer is:

---

detect any glimpse of light and *relay* is not sensitive enough to make the switch $sw_2$ jump off quickly enough then *detect* will be on; otherwise it will stay off.

5. Note that this point of view also corresponds to the traditional definition of states as snapshots of the world.

6. For example, given *detect* is not on in the next state, it can be that either the *light* has never been bright or the *light* may have been bright but the *detect* is not sufficiently sensitive to detect its momentary brightness.

In general, when both latches are up, it is impossible to perform *only* the action of closing the suitcase; one must also concurrently toggle at least one of the latches.

(McCain & Turner, 1997, p. 464, *italic is original.*)

The problem now is how to represent the action of holding the suitcase closed such that it would overcome the above indirect effect caused by the loaded spring. This also suggests another kind of actions whose direct effects are to keep the world unchanged. These actions have usually been formalised by other researchers as fluents, e.g. *holding*. Our main objection to this approach is that agents also need to reason about these actions since they may also require certain preconditions such as the agent is strong enough to hold the object. Moreover, under certain (abnormal) circumstances, agents may also fail to perform such actions. That is, these actions are also subject to the qualification problem discussed in the previous subsection. ☐

**Example 3** Consider the circuit in Figure 2:[7]

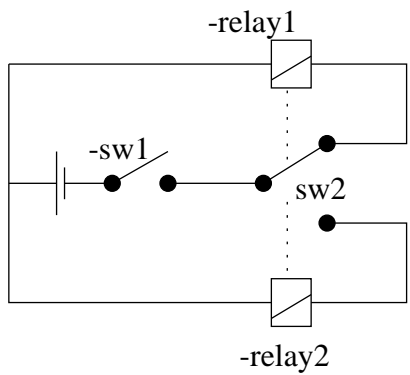

Figure 2: A dynamic domain with a (potentially) infinite sequence of indirect effects

It is quite obvious that after performing the action $flip_1$ whose direct effect is having $sw_1$ closed, the following circular sequence of indirect effects will take place: $\{relay_1, \neg relay_2\} \rightarrow \neg sw_2 \rightarrow \{\neg relay_1, relay_2\} \rightarrow sw_2 \rightarrow \{relay_1, \neg relay_2\}$. This sequence of course would potentially carry on the above sequence of indirect effects indefinitely unless $sw_1$ is flipped open or some device stopped functioning correctly, e.g. when the battery is out of charge. In other words, this action domain requires some action to be inserted in between a series of on going indirect effects which can not be captured by the above representation. Note also that none of the causation-based representations proposed by Lin (1995), McCain and Turner (1995) or Thielscher (1997) is able to deal with the above action domain. ☐

## 1.4 Towards a Solution

To address the problems discussed in the previous sections, we argue that in order to find a uniform solution to these problems one should avoid cryptic formalisms whose consequences can not be seen clearly from the formalisation of the problem domains. As a consequence, we propose a uniform non-monotonic solution to the main problems of reasoning about action. Essentially, when performing commonsense reasoning, the reasoner relies on a number of plausible assumptions, e.g., assuming that an instance of birds flies, or assuming that shooting a turkey with a loaded gun causes it to die, etc. In traditional default reasoning formalisms such as circumscriptive approaches or default logic, these assumptions are made implicit. For example, these are the instances of predicates which are minimised away by circumscription or the implicitly asserted justifications in default rules when they are still consistent with the extension under consideration in default logic. The proposed representation formalism aims at making these assumptions explicit so that an automated reasoner is conscious (at least) about what assumptions it relies on when performing reasoning. Then the reasoner can always manipulate these assumptions independently of each other. It is also the basic idea of assumption-based frameworks which are at heart of Bondarenko, Dung, Kowalski, and Toni's (1997) argumentation-theoretic approach.

---

7. This example is an instance of the so-called "vicious cycles" scenarios, e.g., see (Shanahan, 1999).

We then proceed to consider the ramification problems and domain theories with concurrent and non-deterministic events. Among the major results, we show that our framework captures the essence of the causation-based approaches regarding the ramification problem. Moreover, we also show the expressiveness of our formalism through two examples in which indirect effects also need qualifications and infinite sequence of indirect effects. To the best of our knowledge, none of the existing formalisms are able to cope with these scenarios.

Based on the basic idea of assumption-based frameworks, our approach comprises the following major aspects of representation:

1. We introduce different types of assumptions to render various laws of common sense in dynamic domains. For instance, frame assumptions are introduced to capture the *common sense law of inertia* whilst (two types of) qualification assumptions are introduced to overcome the qualification problem.

2. We introduce a special class of (system-generated) *dummy actions* to allow the explanation problem, i.e. when some actions or events occur outside of the reasoner's knowledge, to be dealt with in a uniform manner.

3. Being based on Bondarenko et al.'s (1997) argumentation-theoretic framework, our approach makes use of the inference rules to represent domain knowledge.

4. Lying at heart of our approach is an argumentation-theoretic semantics, called *plausibility semantics*, which is argued to best render common sense knowledge in dynamic domains. This semantics consists in a particular policy of resolving conflicting assumptions when computing the argumentation to be accepted.

To summarise, in this paper we formalise an expressive representation scheme in order to cope with sophisticated action domains. We believe that such a formalisation sometimes requires certain advanced knowledge to be encoded in a precise and well-engineered way. The representation of action theories proposed in this paper can be considered as the intermediate level between commonsense and scientific knowledge. The expressiveness of the formalism is improved through several independent steps by adding further assumptions into the domain descriptions. This also shows one advantage of our solution: a simple representation can be achieved by simply removing the involved assumptions. This is arguably a desirable feature as the reasoner has the option of either increasing the expressibility of the representation formalism or improving the simplicity and, as a consequence, the efficiency of the reasoning system.

The rest of the paper is organised as follows: Section 2 summarises relevant features of the abstract argumentation framework proposed by Bondarenko et al. (1997), its semantics and concrete instances. In Section 3 we present the syntax and semantics of the basic temporal logic and the extension for reasoning about action. In Section 4 we present our formalisation for reasoning about action based on the argumentation-theoretic approach introduced by Bondarenko et al. (1997). Our approach to reasoning about action, in particular a uniform solution to the frame and the qualification problems, as well as the main results of the paper are presented in Section 5. In Section 6, we show how the proposed formalism is extended to deal with more complex dynamic domains, including those with concurrent and non-deterministic events, and indirect effects. Related work and future research directions are discussed in Section 7. We will defer most proofs of the results presented in the paper to the Appendix.

This paper is an extended version of two earlier conference papers (Vo & Foo, 2001, 2002). The main differences are that in this version all proofs are included, lemmas that are used in the proofs of the theorems are introduced to help the reader more easily comprehend these results, and the presentation has been improved and extended with more examples of the various constructions.

## 2. Defeasible Reasoning by Argumentation

Let a *deductive system* $\langle \mathcal{L}, \mathcal{R} \rangle$ be given, where $\mathcal{L}$ is some formal language with countably many sentences and $\mathcal{R}$ is a set of inference rules. Given a theory $\Gamma \subseteq \mathcal{L}$ and a sentence $\alpha \in \mathcal{L}$, we write $\Gamma \vdash_{\langle \mathcal{L}, \mathcal{R} \rangle} \alpha$ if there is a deduction from $\Gamma$ whose last element is $\alpha$. $Th_{\langle \mathcal{L}, \mathcal{R} \rangle}(\Gamma)$ denotes the set $\{\alpha \in \mathcal{L} \mid \Gamma \vdash_{\langle \mathcal{L}, \mathcal{R} \rangle} \alpha\}$. Since the language $\mathcal{L}$ is generally kept fixed whereas the set of inference rules $\mathcal{R}$ is likely to vary depending on the description of the domain, when there is no possible confusion we will abbreviate $\vdash_{\langle \mathcal{L}, \mathcal{R} \rangle}$ and $Th_{\langle \mathcal{L}, \mathcal{R} \rangle}$ as $\vdash_{\mathcal{R}}$ and $Th_{\mathcal{R}}$, respectively. Thus the classical inference relation $\vdash$ can also be written as $\vdash_{\mathcal{R}_C}$ where $\mathcal{R}_C$ is the set of inference rules of classical propositional logic. Note also that every set of inference rules considered in this paper will be a super set of $\mathcal{R}_C$.

Given a deductive system $\langle \mathcal{L}, \mathcal{R} \rangle$, an *assumption-based framework* with respect to $\langle \mathcal{L}, \mathcal{R} \rangle$ consists of a theory $\Gamma$ representing the current knowledge of the reasoner about the domain, an assumption base $\mathcal{AB}$ and a contrariness operator $^-$, i.e. given an assumption $\delta \in \mathcal{AB}$, $\overline{\delta}$ denotes the *contrary* of $\delta$.

REMARK: The notion of the contrary of an assumption is intended to generalise the classical negation $\neg \delta$. Note that in general assumptions may be constructed by special operators (e.g. *negation-as-failure* in the case of logic programming, or the modal operator $L$ in the case of autoepistemic logic, Moore, 1985), thus the contrariness operator must also be sufficiently general.

The hardest part in reasoning with assumption-based frameworks is computing the set of assumptions to augment the given theory $\Gamma$. In an argumentation-theoretic approach, this is realised by the attack relation. To determine which assumptions to be accepted, assumptions are put together to form arguments. The assumptions behind the best arguments are considered to be acceptable. Several semantics for best arguments are presented by Bondarenko et al. (1997) based on the notions of attack: Given an assumption-based framework $\langle \Gamma, \mathcal{AB}, {}^{-} \rangle$ and an assumption set $\Delta \subseteq \mathcal{AB}$:

- $\Delta$ *attacks an assumption* $\delta \in \mathcal{AB}$ iff $\overline{\delta} \in Th(\Gamma \cup \Delta)$.

- $\Delta$ *attacks an assumption set* $\Delta' \in \mathcal{AB}$ iff $\Delta$ attacks some assumption $\delta \in \Delta'$.

- $\Delta$ is *closed* iff $\Delta = \mathcal{AB} \cap Th(\Gamma \cup \Delta)$.

- $\Delta$ is *conflict-free* iff there does not exist any $\delta \in \mathcal{AB}$ such that $\Gamma \cup \Delta \vdash_{\mathcal{R}} \delta, \overline{\delta}$.

Assumption-based frameworks in which assumption sets are always closed are referred to as *flat*. In a flat assumption-based framework, the conflict-free property of a set of assumptions $\Delta$ is equivalent to the property that $\Delta$ does not attack itself. The major argumentation-theoretic semantics defined by Bondarenko et al. (1997) for assumption-based frameworks include:

- *Stability semantics*: an assumption set $\Delta \subseteq \mathcal{AB}$ is *stable* iff

  1. $\Delta$ is closed,

  2. $\Delta$ does not attack itself, and

  3. $\Delta$ attacks each assumption $\delta \notin \Delta$.

  Bondarenko et al. (1997) show that the above stability semantics corresponds to the standard semantics of extensions of Theorist (Poole, 1988), minimal models of (many cases of) circumscription (McCarthy, 1980, 1986), extensions of Default Logic (Reiter, 1980), stable expansions of Autoepistemic Logic (Moore, 1985), and stable models of logic programming. In other words, from a complexity-theoretic perspective, any approach based on the existing formalisms to default reasoning can be rendered in a corresponding assumption-based argumentation framework with no loss in terms of computational complexity.

- *Admissibility and Preferability semantics*: Bondarenko et al. (1997) go further to extend these existing formalisms by generalising the semantics of admissible and preferred arguments which were originally proposed for logic programming only. The new semantics are defined in terms of "admissible" and "preferred" sets of assumptions/extensions. An assumption set $\Delta \subseteq \mathcal{AB}$ is *admissible* iff

  1. $\Delta$ is closed,

  2. $\Delta$ does not attack itself, and

  3. for all closed sets of assumptions $\Delta' \subseteq \mathcal{AB}$ if $\Delta'$ attacks $\Delta$ then $\Delta$ attacks $\Delta'$.

Maximal (with respect to set inclusion) admissible assumption sets are called *preferred*.

Throughout this paper assumptions are expressed in terms of usual propositions. Thus, we will replace the notion of contrariness $^{-}$ in Bondarenko et al.'s (1997) system with the classical negation $\neg$ and omit it from the specification of assumption-based frameworks. That is, an assumption-based framework $\langle \Gamma, \mathcal{AB} \rangle$ consists of a theory $\Gamma \subseteq \mathcal{L}$, and the assumption base $\mathcal{AB}$ which contains the assumptions to be used in the reasoning.

## 3. Domain Descriptions

We introduce a propositional action description language based on a more comprehensive representation formalism proposed by Sandewall (1994). In particular, we extend Draken-gren and Bjä reland's (1999) language so that it is possible to describe narratives in our framework.

### 3.1 Syntax

Following Sandewall (1994), the underlying representation of time is a (*discrete*) *time structure* $\mathbf{T} = \langle \mathbb{T}, <, +, - \rangle$ consisting of

- a *time domain* $\mathbb{T}$ whose members are called *time points* which are integers in this paper (except in a later part of the paper where the distinction will be made explicit);

- $<, +, -$ are as usual for integers.

Given a time structure $\mathbf{T} = \langle \mathbb{T}, <, +, - \rangle$, a *signature* $\sigma$ with respect to $\mathbf{T}$ is a tuple $= \langle \mathcal{T}, \mathcal{F}, \mathcal{A} \rangle$, where $\mathcal{T}$ is a set of countably infinitely many *time-point variables*, $\mathcal{F}$ is a set of *propositional fluent names*, and $\mathcal{A}$ is a set of *action names*. Since the time structure $\mathbf{T}$ is fixed in the rest of this paper, $\mathbf{T}$ will be taken implicitly whenever a signature is introduced. We assume that all sets in $\sigma$ are countable. We denote $\mathcal{F}^* = \{[\neg]f \mid f \in \mathcal{F}\}$.[8] A member of $\mathcal{F}^*$ is a *fluent literal*. Moreover, $\mathcal{A} = \mathcal{A}_0 \cup \mathcal{DA}$, where $\mathcal{A}_0$ is the set of domain dependent action names, called *basic actions*, e.g. *load*, *shoot*, etc. and $\mathcal{DA} = \{da_l \mid l \in \mathcal{F}^*\}$ is the set of *dummy actions*. As will be explained later in this paper, our solution to the problems of reasoning about action is based on the basic guideline of attributing changes to events. Given the reasoner's ignorance about certain events that bring about changes in the world, the dummy actions are to be used to make up for these gaps in the reasoner's belief state. That is why we need to associate the dummy actions with the fluent literals from $\mathcal{F}^*$.

For each fluent literal $l \in \mathcal{F}^*$, we introduce the following two symbols: $AQ_l$, and $FA_l$:

- $AQ_l$ is associated with the assumed qualifications upon the preconditions of an action regarding the fluent literal $l$. Essentially, $AQ_l$ when used in the description of the dynamics of an action $\alpha$ allows the reasoner to describe only the main preconditions of $\alpha$ (with regards to the fluent literal $l$) while leaving other possible (but less probable) qualifications to be rendered by a single assumption $AQ_l$.

- $FA_l$ is associated with the frame assumptions regarding $l$. $FA_l$, when coupled with a particular *frame inference rule*, allows the reasoner to infer that the fluent literal $l$ continues to hold in future time points unless there is a reason that defeats $FA_l$.

---

8. The notation $[\neg]$ means that the formula following it may, or may not, be negated.

Given a set of fluent literals $\Gamma \subseteq \mathcal{F}^*$, we denote $FA_\Gamma \stackrel{def}{=} \{FA_l \mid l \in \Gamma\}$ and $AQ_\Gamma \stackrel{def}{=} \{AQ_l \mid l \in \Gamma\}$.

A *time-point expression* is one of the following:

- a member of $\mathbb{T}$,

- a time-point variable in $\mathcal{T}$,

- an expression formed from time-point expressions using $+$ and $-$. For convenience, we will also write $\tau^+$ and $\tau^-$ instead of $\tau + 1$ and $\tau - 1$, respectively.

We denote the set of time-point expressions by $\mathcal{TE}$.

**Definition 3.1** Let a signature $\sigma = \langle \mathcal{T}, \mathcal{F}, \mathcal{A} \rangle$ be given and $\tau, \upsilon \in \mathcal{TE}$, $f \in \mathcal{F}$, $\alpha \in \mathcal{A}$, $R \in \{=, <\}$, $\otimes \in \{\wedge, \vee, \rightarrow, \leftrightarrow\}$. Define the *basic (domain description) language* $\Lambda$ over $\sigma$ by:

$\quad \Lambda_0 ::= \mathbf{true} \mid \mathbf{false} \mid \mathbf{f} \mid \tau\mathbf{R}\upsilon \mid \neg\boldsymbol{\Lambda_0} \mid \boldsymbol{\Lambda_0} \otimes \boldsymbol{\Lambda_0} \mid [\tau]\boldsymbol{\Lambda_0},$

$\quad \Lambda ::= \Lambda_0 \mid [\tau, \upsilon]\alpha \mid \neg\Lambda \mid \Lambda \otimes \Lambda$

and the *assumption base* $\mathcal{AB}$ by:

$\quad \mathcal{AB} = \mathcal{AB}_{AQ} \cup \mathcal{AB}_{FA}$, where

$\quad \mathcal{AB}_{AQ} = \{[\tau, \upsilon]AQ_l \mid \tau, \upsilon \in \mathcal{TE} \text{ and } l \in \mathcal{F}^*\}$, and

$\quad \mathcal{AB}_{FA} = \{[\tau]FA_l \mid \tau \in \mathcal{TE} \text{ and } l \in \mathcal{F}^*\}$.

The *domain description language* $\mathcal{L}_D$ (over $\sigma$) is defined: $\mathcal{L}_D = \Lambda \cup \mathcal{AB}$.[9]

$[\tau, \upsilon]\alpha$ means the action $\alpha$ has a duration corresponding to the interval $[\tau, \upsilon]$. $[\tau, \upsilon]AQ_l$ means the fluent literal $l$ is assumed to be qualified to hold by the end of the interval $[\tau, \upsilon]$. $[\tau]FA_l$ means the fluent literal $l$ is assumed by default to persist from the time point $\tau$ to the next, i.e. *the principle of inertia*. Notice the difference between $[\tau]FA_{l_1}$ and $[\tau]l_2$ for some fluent literals $l_1, l_2 \in \mathcal{F}^*$. $[\tau]l_2$ indicates that the fluent literal $l_2$ holds at $\tau$ while $[\tau]FA_{l_1}$ indicates that the fact that the fluent literal $l_1$ persists during the interval $[\tau, \tau^+]$ is true.

For example, in the blocks world domain, to say that block $A$ is on block $B$ at the time point 2, we write: $[2]on(A, B)$; or, to say that an action *pickup* the block $A$ occurs between time points $t_1 + 3$ and $t_2 - 1$ and that the relation $<$ holds between $t_1 + 5$ and $t_2$, we write $[t_1 + 3, t_2 - 1]pickup(A) \wedge (t_1 + 5 < t_2)$.

A formula that does not contain any connectives (i.e. $\wedge, \vee, \rightarrow, \leftrightarrow, \neg$, and $[.]$) is *atomic*. If $\gamma$ is atomic and $\tau \in \mathcal{TE}$, then the formulas $\gamma$, $[\tau]\gamma$, $\neg\gamma$, $\neg[\tau]\gamma$, and $[\tau]\neg\gamma$ are *literals*.

Let $\gamma$ be a formula. A fluent $f \in \mathcal{F}$ occurs *free* in $\gamma$ iff it does not occur within the scope of a $[\tau]$ expression in $\gamma$. $\tau \in \mathcal{TE}$ *binds* $f$ in $\gamma$ if a formula $[\tau]\psi$ occurs as a subformula of $\gamma$, and $f$ is free in $\psi$. If no fluent occurs free in $\gamma$, $\gamma$ is *closed*. If $\gamma$ does not contain any occurrence of $[\tau]$ for any $\tau \in \mathcal{TE}$, then $\gamma$ is *propositional*.

---

9. It would be more precise to denote the domain description language over $\sigma$ by $\mathcal{L}_\sigma$. However, as the signature is usually clear from the context and in order to avoid the mention of $\sigma$ every time we have to formalise something with the domain description language, we choose to denote it by $\mathcal{L}_D$.

### 3.2 Semantics

**Definition 3.2** Let $\sigma = \langle \mathcal{T}, \mathcal{F}, \mathcal{A} \rangle$ be a signature. A *state* over $\sigma$ is a function from $\mathcal{F}$ to the set $\{\textbf{true}, \textbf{false}\}$ of truth values. A *history* over $\sigma$ is a function $h$ from $\mathbb{T}$ to the set of states. A *valuation* is a function $\phi$ from $\mathcal{TE}$ to $\mathbb{T}$. A *narrative assignment* is a function $\eta$ from $\mathbb{T} \times \mathcal{A} \times \mathbb{T}$ to the set $\{\textbf{true}, \textbf{false}\}$. In addition, we define $\varepsilon_q : \mathbb{T} \times \mathbb{AQ}_{\mathcal{F}^*} \times \mathbb{T} \to \{\textbf{true}, \textbf{false}\}$ and $\varepsilon_f : \mathbb{T} \times \mathbb{FA}_{\mathcal{F}^*} \to \{\textbf{true}, \textbf{false}\}$. An *interpretation* over $\sigma$ is a tuple $\langle h, \phi, \eta, \varepsilon_q, \varepsilon_f \rangle$ where $h$ is a history, $\phi$ is a valuation, $\eta$ is a narrative assignment and $\varepsilon_q, \varepsilon_f$ are defined as above.

**Example 4** Consider Hanks and McDermott's (1987) Yale Shooting Problem (YSP) : There are three possible actions: *load* (the gun), *wait*, and *shoot* (the victim with the gun). Normally, waiting does not cause any change in the world, but shooting leads to the victim's death, provided that, of course, the gun is loaded. Assume that all three actions are performed, in the given order.

We define the signature $\sigma_{\text{YSP}}$ to be a tuple $\langle \{t, t_1, t_2, \ldots, u, u_1, u_2, \ldots\}, \{loaded, alive\}, \{load, wait, shoot\} \rangle$. Then the Yale Shooting problem can be formulated in the domain description language $\mathcal{L}_{\text{YSP}}$ as the following theory: $\Gamma_{\text{YSP},0} = \{[0]alive,\ [0,1]load,\ [1,2]wait,\ [2,3]shoot\}$.

The following two histories $h_1$ and $h_2$ are corresponding to the well-known models in the literature of reasoning about action: $h_1$ to the intended model and $h_2$ to the anomalous model most frameworks would produce.

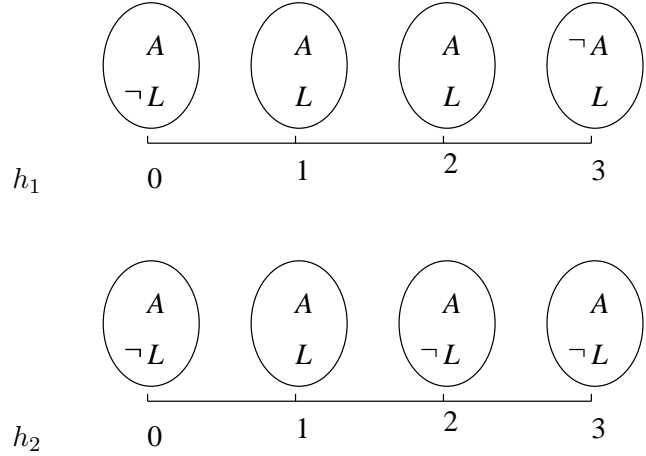

Figure 3: The two histories for the YSP action description.

Each oval in Figure 3 represents a state over $\sigma_{\text{YSP}}$. A narrative assignment complying with the above action description would map the three tuples $(0, load, 1), (1, wait, 2)$, and $(2, shoot, 3)$ to **true** and other tuples to **false** (relative to the assumption that *'normally, given any action and any time point, there is no instance of that action at that time point unless specified otherwise'*).

**Definition 3.3** Let $\gamma, \delta \in \Lambda$ and $I = \langle h, \phi, \eta, \varepsilon_q, \varepsilon_f \rangle$ an interpretation. Assume $\tau, v \in \mathcal{TE}$, $f \in \mathcal{F}$, $A \in \mathcal{A}$, $R \in \{=, <\}$, $l \in \mathcal{F}^*$, $\otimes \in \{\wedge, \vee, \to, \leftrightarrow\}$, and $\chi \in \{\textbf{true}, \textbf{false}\}$. Define the truth value of $\gamma$ in $I$ for a time point $t \in \mathbb{T}$, denoted $I(\gamma, t)$ as follows:

$$I(\chi, t) = \chi \qquad\qquad\qquad I(f, t) = h(t)(f)$$
$$I([\tau, v]A, t) = \eta(\phi(\tau), A, \phi(v)) \qquad I([\tau, v]AQ_l, t) = \varepsilon_q(\phi(\tau), AQ_l, \phi(v))$$
$$I([\tau]FA_l, t) = \varepsilon_f(\phi(\tau), FA_l) \qquad I(\tau Rv, t) = \phi(\tau)R\phi(v)$$
$$I(\neg\gamma, t) = \neg I(\gamma, t) \qquad\qquad I(\gamma \otimes \delta, t) = I(\gamma, t) \otimes I(\delta, t)$$
$$I([\tau]\gamma, t) = I(\gamma, \phi(\tau))$$

Two formulas $\gamma$ and $\delta$ are equivalent iff $I(\gamma, t) = I(\delta, t)$ for all $I$ and $t$. An interpretation $I$ is a *model* of a set $\Gamma \subseteq \Lambda$ of formulas, denoted $I \models \Gamma$, iff $I(\gamma, t) = \textbf{true}$ for every $t \in \mathbb{T}$ and $\gamma \in \Gamma$. A formula $\gamma \in \Lambda$ is *entailed* by a set $\Gamma \subseteq \Lambda$ of formulas, denoted $\Gamma \models \gamma$, iff $\gamma$ is true in all models of $\Gamma$.

**Definition 3.4** Let $I = \langle h, \phi, \eta, \varepsilon_q, \varepsilon_f \rangle$ be an interpretation.

1. The set $Occ^I = \{(t, A, u) \in \mathbb{T} \times \mathcal{A} \times \mathbb{T} \mid \eta(t, A, u) = \textbf{true}\}$ is called *action occurrence denotation* of $I$.

2. The set $FA^I = \{(t, FA_l) \in \mathbb{T} \times \mathbb{FA}_{\mathcal{F}^*} \mid \varepsilon_f(t, FA_l) = \textbf{true}\}$ is called *FA-denotation* of $I$.

3. The set $AQ^I = \{(t, AQ_l, u) \in \mathbb{T} \times \mathbb{AQ}_{\mathcal{F}^*} \times \mathbb{T} \mid \varepsilon_q(t, AQ_l, u) = \textbf{true}\}$ is called *AQ-denotation* of $I$.

# 4. Representing Dynamic Domains in the Argumentation-Theoretic Approach

We now proceed to showing an assumption-based framework for representing dynamic domains. We subsequently introduce a uniform framework for solving the frame and the qualification problems based on the argumentation-theoretic approach. General solutions for the frame and the qualification problems can be obtained by computing plausible sets of assumptions which guarantee that extensions computed from these sets of assumptions will be consistent when the given theory is consistent. We now introduce some additional notations: Given an inference rule $r \in \mathcal{R}$, we denote by $prem(r)$ and $cons(r)$ the premise and the consequence of rule $r$, respectively.

**Definition 4.1** A deductive system $\langle \mathcal{L}_D, \mathcal{R} \rangle$ is *well-defined* iff for each subset $S \subseteq \mathcal{R}$, if the set $\bigcup_{r \in S} prem(r)$ is consistent then the set $CONS(S) = \{cons(r) \mid r \in S\}$ is also consistent.

Henceforth, we will assume that deductive systems are well-defined. Being formalised in terms of the argumentation-theoretic approach, the representation requires an extended notion of consistency.

**Definition 4.2** Let $\langle \mathcal{L}_D, \mathcal{R} \rangle$ be a deductive system,

(i) a set of sentences $\Gamma \subseteq \mathcal{L}_D$ is $\mathcal{R}$-*consistent* iff $\Gamma \not\vdash_{\mathcal{R}} \textbf{false}$;

(ii) an assumption-based framework $\langle \Gamma, \mathcal{AB} \rangle$ with respect to $\langle \mathcal{L}_D, \mathcal{R} \rangle$ is *consistent* iff $\Gamma$ is $\mathcal{R}$-consistent.

REMARK: Observe that even when $\langle \mathcal{L}_D, \mathcal{R} \rangle$ is a well-defined deductive system, consistency is not equivalent to $\mathcal{R}$-consistency. For instance, let $\mathcal{R} = \{\frac{b}{\neg a}\}$, the (logically) consistent theory $\Gamma = \{a, b\}$ is not $\mathcal{R}$-consistent.

**Example 4** (*continued*) Returning to the Yale Shooting problem, the following inference rules describe the actions of this domain:

$$\frac{[\tau, v]load}{[v]loaded \wedge \neg[\tau]FA_{\neg loaded}} \tag{5}$$

$$\frac{[\tau, v]shoot, [\tau]loaded}{\neg[v]alive \wedge \neg[\tau]FA_{alive}} \tag{6}$$

$$\frac{[\tau]loaded, [\tau]FA_{loaded}}{[\tau^+]loaded} \tag{7}$$

$$\frac{[\tau]alive, [\tau]FA_{alive}}{[\tau^+]alive} \tag{8}$$

$$\frac{\neg[\tau]loaded, [\tau]FA_{\neg loaded}}{\neg[\tau^+]loaded} \tag{9}$$

$$\frac{\neg[\tau]alive, [\tau]FA_{\neg alive}}{\neg[\tau^+]alive} \tag{10}$$

Rules (5) and (6) represent the descriptions of the actions *load* and *shoot*, respectively. Action *wait* does not cause any effect to the world, so there is no need to describe it. Other rules render the common sense law of inertia: *"at any time point, a fluent literal presumably persists to the next time point."*

Most argumentation-theoretic semantics, e.g. stability, admissibility, preferability, complete, well-founded semantics, etc. (Bondarenko et al., 1997) are based on the notion of attack. However, to reason about problem domains with incomplete information, especially action domains, this notion alone may not be sufficient as we may not always be able to construct explicit arguments to defeat unsound assumptions. For example, consider the Yale Shooting Problem: By observing that a turkey is shot with a loaded gun at time point 1, the reasoner infers plausibly that the turkey is dead at time point 2 using the assumption that the action shoot is qualified to bring about the effect of killing the victim. However, at time point 2, the reasoner could observe that the turkey is still alive. Existing solutions to the frame problem, e.g. Reiter's (1991), Thielscher's (1997), Castilho et al.'s (1999), etc. fail to deal with such a surprise since they allow a contradiction to be derived. Observe that the reasoner does not have any explicit reason to defeat the above qualification assumption, i.e. she is not aware of any cause that prevents the application of this qualification assumption. She only knows that it is not acceptable in this case by common sense. To formalise such phenomena, we introduce the notion of rejected assumptions.

**Definition 4.3** Given an assumption-based framework $\langle \Gamma, \mathcal{AB} \rangle$, a set of assumptions $\Delta \subseteq \mathcal{AB}$ *rejects* an assumption $\delta \in \mathcal{AB}$ iff

   (a) $\Delta$ is conflict-free, and

   (b) $\Delta \cup \{\delta\}$ attacks itself.

For instance, in example 4, the set of assumptions $\Delta_1 = \{[0]FA_{alive}, [1]FA_{alive}, [1]FA_{loaded}\}$ attacks the assumption $[2]FA_{alive}.$[10] Moreover, relative to the given action description, any set of assumptions attacks the assumption $[0]FA_{\neg loaded}$. On the other hand, the set of assumptions $\Delta_2 = \{[0]FA_{alive}, [1]FA_{alive}, [2]FA_{alive}\}$ rejects the assumption $[1]FA_{loaded}$ but $\Delta_2$ does **not** attack it.

**Observation 1** *Given an assumption-based framework $\langle \Gamma, \mathcal{AB} \rangle$ and a conflict-free set of assumptions $\Delta \subseteq \mathcal{AB}$, if $\Delta$ attacks an assumption $\delta \notin \Delta$ then $\Delta$ rejects $\delta$.*

Then, why do we not generalise the contrariness notion of an assumption so that it would be general enough to account for all rejected assumptions? The reason is because we want to isolate the set of assumptions that are rejected but are not attacked as part of our solution to the frame problem.

**Definition 4.4** Given an assumption-based framework $\langle \Gamma, \mathcal{AB} \rangle$, a set of assumptions $\Delta \subseteq \mathcal{AB}$ *leniently rejects* an assumption $\delta \in \mathcal{AB}$ iff

   (a) $\Delta$ rejects $\delta$, and

   (b) $\Delta$ does not attack $\delta$.

We denote $Lr(\Delta) \overset{def}{=} \{\delta \in \mathcal{AB} \mid \delta$ is leniently rejected by $\Delta\}$.

To show that our solution provides an intuitive account for problems of reasoning about action, several scenarios should be considered. These include the projection problem, the most basic form of the frame problem, whose typical example is the infamous YSP. Another scenario concerns with the explanation problem which are usually discussed with the Stolen Car Problem (Kautz, 1986) and the Stanford Murder Mystery (Baker, 1989). We will first provide an informal discussion of our approach through these examples.

In the present solution, the frame assumptions are the essence of the principle of inertia,and their role in the argumentation approach is illustrated below by the Yale Shooting Problem. In this formulation we intentionally ignore the qualification problem (it is addressed in the next section) to highlight how the frame problem is solved. We now reconsider the well-worn example YSP to motivate our approach to the frame problem.

**Example 4** (*continued*) Given the theory $\Gamma_{\text{YSP}}$, the argumentation-theoretic approach will yield the following preferred set of assumptions (Bondarenko et al., 1997):

   $\{[t]FA_l \mid t \in \mathbb{T}$ and $l \in \{loaded, alive, \neg loaded, \neg alive\}\} \setminus \{[0]FA_{\neg loaded}, [2]FA_{alive}\},$
which corresponds to the intended model of this scenario in which the gun remains loaded at time point 2 and the victim is not alive at time point 3.

This extension is also the stable extension and well-founded semantics (Bondarenko et al., 1997) of the given theory under the argumentation-theoretic approach. Note that in case one would like to be uncertain about whether the gun is still loaded after the shooting action, one just simply needs to add an axiom: $[\tau, v]shoot \rightarrow \neg[\tau]FA_{loaded}$ to dictate that the persistence of the fluent *loaded* after the action shooting is not guaranteed. In that case, we can still derive that $[\tau]loaded$ for $\tau = 1, 2$, but we can no longer give a definite assertion about $[\tau]loaded$ for $\tau \geq 3$.

---

10. In fact, any set of assumptions containing the assumption $[1]FA_{loaded}$ would attack $[2]FA_{alive}$.

As the above formalisation of YSP resembles that using default logic, it may be surprising that the problem of unintended models pointed out by Hanks and McDermott (1987) for circumscription, default logic, autoepistemic logic does not happen here. The principal reason is the interaction of the inference rules and the notion of attack in the argumentation-theoretic framework, which invalidates undesired assumptions. Notice that even if $\neg[2]loaded$ can be (magically) derived, it cannot lead to $\neg[1]FA_{loaded}$. Therefore, the set of assumptions corresponding to this case does not satisfy the conditions of preferred set of assumptions, thus ruling out this unintended model. This shows one of the important features of assumption-based frameworks on its capability of making explicit the assumptions used by the reasoner during the course of inference. Recall that defaults' justifications are accepted as long as they are consistent with some extension for credulous semantics or all extensions for skeptical semantics (thus the name consistency-based approach.) In light of the inertia principle, it's considered to be abnormal if a fluent does not persist from a state to the next state. To minimise the abnormality, (normal) default rules are introduced to express the fact that if it's consistent to believe that there is no abnormality with respect to a fluent $f$ and an action $a$ in a situation $s$ then assert that. But then we would fail to distinguish between the abnormalities brought about by reasonable causes and those unintuitively generated to make them consistent with some possible extension. The latter is of course corresponding to the anomalous models. By using explicit assumptions, not only consistency is maintained (by preventing the accepted assumptions from attacking themselves) but each rejection of assumptions must also be justified by the known facts from the given theory.

**Discussion:**

1. Turner (1997) showed that an alternative representation of the YSP in default logic can help solve the issue of anomalous models introduced by Hanks and McDermott's (1987) representation. Turner formulates the Yale Shooting scenario as follows:

$$\frac{\neg Holds(Alive, S_0)}{\textbf{False}} \tag{11}$$

$$\frac{\textbf{True}}{Holds(Loaded, Result(Load, s))} \tag{12}$$

$$\frac{Holds(Loaded, s)}{\neg Holds(Alive, Result(Shoot, s))} \tag{13}$$

$$\frac{:\ Holds(f, S_0)}{Holds(f, S_0)} \tag{14}$$

$$\frac{:\ \neg Holds(f, S_0)}{\neg Holds(f, S_0)} \tag{15}$$

$$\frac{Holds(f, s):\ Holds(f, Result(a, s))}{Holds(f, Result(a, s))} \tag{16}$$

$$\frac{\neg Holds(f, s):\ \neg Holds(f, Result(a, s))}{\neg Holds(f, Result(a, s))} \tag{17}$$

Notice that Turner also uses the inference rules to block the "backward" reasoning that generates the anomalous models of the Yale Shooting scenario. However, this also means that all kinds of useful backward reasoning will also be blocked. In other words, Turner's formulation fails to deal with "surprising" observations about states at later time points. As a consequence, Turner's formulation only works when the domain is restricted to qualification-free. As soon as the action descriptions, e.g. the one for the shoot action in YSP, need to rely on some default justifications, e.g. qualification assumptions, Turner's formulation would also encounter the problem of undesirable extensions. Our approach offers solutions to both of the above issues.

2. After Hanks and McDermott's (1987) seminal paper in which early approaches to the frame problems were exposed, besides new attempts to solve the frame problem, Sandewall (1994) should be accredited as the first who tried to approach the problems of reasoning about action in a systematic way. As part of this effort, he also examines the reason behind the failure of early approaches to the frame problem. As discussed by Sandewall, early approaches to reasoning about action while attempting to formulate the inertia principle have made the common mistake of making changes the abnormality regarding this principle but failing to distinguish between normal changes triggered by actions and anomalous changes. This important insight turns out to be a consequence of a much more general law for reasoning about dynamic domains discovered by researchers in the community in pursuit of solutions to various problems of reasoning about actions: *"action dynamics are causality-based."* It is this principle that underpins most solutions to the problems of reasoning about action. The anomalous models that arise in early approaches to the frame problem discovered by Hanks and McDermott (1987) or to the qualification problem as discussed by Thielscher (2001) are those in which the *causes* of abnormalities are not present. On the other hand, regarding the ramification problem, given a domain constraint involving a number of fluents, it's important to know which of these fluents are the causes influencing the other fluents, i.e. the *causality direction* between the involved fluents.

In light of the above analysis, Turner's (1997) approach appears to be rather ad hoc. Note that Turner's solution to the problem of anomalous models is to block "backward reasoning" by the use of inference rules without any motivation on why backward reasoning is a bad thing. While his approach appears to share with solutions based on *chronological ignorance* (which will be discussed in more details in Section 7) the notion of directedness: By minimizing chronologically or blocking backward reasoning, one tends to minimize causes rather than effects. However, a more systematic approach to various problems of reasoning about action is still very much desired.

Nevertheless, while the preferability semantics copes successfully with the YSP, it can not properly account for the explanation problem, e.g. the Stanford Murder Mystery (Baker, 1989), the Stolen Car Problem (Kautz, 1986). The subtlety lies in the derivation of the contrary of the frame assumptions. The contrary of a frame assumption is derived only when both the occurrence of the event that brings about the change (absent in the Stolen Car Problem) and the preconditions required to be satisfied for the change to actually take place (absent in the Stanford Murder Mystery) are explicitly derivable. This is where the notion of (leniently) rejected assumptions is called into service.

**Definition 4.5** Given an assumption-based framework $FA = \langle \Gamma, \mathcal{AB} \rangle$, a set of assumptions $\Delta \subseteq \mathcal{AB}$ is *presumable wrt $FA$* iff

(a) $\Delta = \{\delta \in \mathcal{AB} \mid \Gamma \cup \Delta \vdash_{\mathcal{R}} \delta\}$ (in the terms given in Bondarenko et al., 1997, $\Delta$ is closed),

(b) $\Delta$ does not attack itself, and

(c) for each assumption $\delta \notin \Delta$, $\delta$ is rejected by $\Delta$.

**Definition 4.6** Given an assumption-based framework $FA = \langle \Gamma, \mathcal{AB} \rangle$, a set of assumptions $\Delta \subseteq \mathcal{AB}$ is *plausible wrt $FA$* iff

(a) $\Delta$ is presumable, and

(b) there exists no $\Delta' \subseteq \mathcal{AB}$ such that $\Delta'$ is presumable and $Lr(\Delta') \subset Lr(\Delta)$.

We now proceed to formalising action theories in our framework.

**Definition 4.7** Let $\sigma = \langle \mathcal{T}, \mathcal{F}, \mathcal{A} \rangle$ be a signature. Assume $\tau, \upsilon \in \mathcal{TE}$, $\alpha \in \mathcal{A}$, $\Phi \subseteq \Lambda$, and $l \in \mathcal{F}^*$. A *domain description $D$* (over $\sigma$) is a tuple $\langle \mathcal{L}_D, \mathcal{R}, \mathcal{AB}, \Gamma \rangle$, where:

1. $\mathcal{L}_D$ is the domain description language and $\mathcal{AB}$ is an assumption base over $\sigma$;

2. $\mathcal{R} = \mathcal{R}_C \cup \mathcal{R}_F \cup \mathcal{R}_A \cup \mathcal{R}_Q$, where

   (a) $\mathcal{R}_C$ is the set of inference rules of (classical) propositional logic;

   (b) $\mathcal{R}_F$ is the set of *frame-based inference rules* of the form: $\dfrac{[\tau]l, [\tau]FA_l}{[\tau^+]l}$, i.e. those that represent the frame axioms in terms of inference rules;

   (c) $\mathcal{R}_A$ is the set of *action descriptions* which are inference rules of the form: $\dfrac{\Phi, [\tau, \upsilon]\alpha, [\tau, \upsilon]AQ_l}{[\upsilon]l \wedge \neg[\tau]FA_{\neg l}}$, i.e. those that represent the conditions for the action $\alpha$ to bring about $l$; and

   (d) $\mathcal{R}_Q$ is the set of *qualification-based inference rules* of the form: $\dfrac{\Phi}{\neg[\tau, \upsilon]AQ_l}$, i.e. those that represent the qualifications regarding the fluent literal $l$.

3. The theory $\Gamma \subseteq \Lambda$.

Given a set of assumptions $\Delta$, we denote $\Delta_{FA} = \Delta \cap \mathcal{AB}_{FA}$ and $\Delta_{AQ} = \Delta \cap \mathcal{AB}_{AQ}$.

**Observation 2** Let $D = \langle \mathcal{L}_D, \mathcal{R}, \mathcal{AB}, \Gamma \rangle$ be a domain description, for each set of assumptions $\Delta \subseteq \mathcal{AB}$, either $\Delta$ is closed or $\Delta$ attacks itself.

## 5. Reasoning about Action: The Frame and the Qualification Problems

In general, we adopt the following guidelines in seeking a uniform solution to the problems of reasoning about action:

- The derived pieces of information do not conflict with the given facts;

- Occurrences of events are minimised; and

- The inertia of fluents is maximised though the minimality of the event occurrences will be of higher priority.

Aside from the trivial case of occurrences of actions causing the frame assumptions to be rejected, two aspects of actions can be distinguished:

1. An action happens but the change it is supposed to cause does not take place. We call this *expectation failure* and this is more or less the qualification problem; and

2. No actions that are known to have happened and caused a change but the change did take place. We call this *surprise* and this is usually known as the explanation problem.

The following assumption represents our underlying intuition behind reasoning about action formalisms.

**Assumption 1** *Intuitive models contain minimal (with respect to set inclusion) sets of surprises.*

Now we introduce some model-theoretic counterpart notions of the assumption-based notions presented above.

**Definition 5.1** Let $\sigma = \langle \mathcal{T}, \mathcal{F}, \mathcal{A} \rangle$ be a signature and $D = \langle \mathcal{L}_D, \mathcal{R}, \mathcal{AB}, \Gamma \rangle$ a domain description over $\sigma$. An interpretation $I = \langle h, \phi, \eta, \varepsilon_q, \varepsilon_f \rangle$ is a *model* of $D$ iff

1. $I$ is a model of $\Gamma$;

2. for each $r \in \mathcal{R}$, if $I \models prem(r)$ then $I \models cons(r)$.

The following definition captures one of several aspects of the (model-theoretic) solution of the frame problem. This aspect is known as the action-oriented frame problem in Lin and Shoham's (1995) terms. The proposed minimisation policy formalises the intuition that change does not happen by itself but is caused by some kind of event. Thus, for each fluent, if its value is changed between two timepoints $\tau$ and $\upsilon$, (at least) an occurrence of some event must end at $\upsilon$ that brings about that change.

**Definition 5.2** Let $D = \langle \mathcal{L}_D, \mathcal{R}, \mathcal{AB}, \Gamma \rangle$ be a domain description and $I$ a model of $D$. $I$ is a *coherent model* of $D$ iff

1. for each basic action $\alpha \in \mathcal{A}_0$ and $\tau, \upsilon \in \mathcal{TE}$, if $I \models [\tau, \upsilon]\alpha$ then $\Gamma \models [\tau, \upsilon]\alpha$ ; and

2. for each $l \in \mathcal{F}^*$ and $t \in \mathbb{T}$, if $I \models [t]l \land \neg[t^+]l$ then either

   (a) there are $\alpha \in \mathcal{A}_0$ and $s \in \mathbb{T}$ such that $r = \dfrac{\Phi, [\tau_1, \tau_2]\alpha, [\tau_1, \tau_2]AQ_{\neg l}}{\neg[\tau_2]l \land \neg[\tau_1]FA_l} \in \mathcal{R}$ and

   $I \models prem(r)[\tau_1/s, \tau_2/t^+]$,[11] or

   (b) $I \models [t, t^+]da_{\neg l}$

Thus, in a coherent model: (i) all satisfiable basic actions must follow from the given theory, and (ii) all changes are attributable to events of one kind or another.

Given an interpretation $I$, we want to extract the sets of assumptions satisfiable in $I$.

**Definition 5.3** Let $\sigma = \langle \mathcal{T}, \mathcal{F}, \mathcal{A} \rangle$ be a signature and $I$ an interpretation over $\sigma$. The set of frame assumptions satisfiable in $I$, denoted $\Delta_{FA}^I$, is defined as follows:

$$\Delta_{FA}^I = \{ [t]FA_l \mid (t, FA_l) \in FA^I \}$$

and the set of qualification assumptions satisfiable in $I$, denoted $\Delta_{AQ}^I$, is :

$$\Delta_{AQ}^I = \{ [t_1, t_2]AQ_l \mid (t_1, AQ_l, t_2) \in AQ^I \}$$

We also write $\Delta_{QF}^I = \Delta_{AQ}^I \cup \Delta_{FA}^I$.

Conversely, given a theory $\Gamma$ and a set of assumptions $\Delta$, a reasoner can also construct his models about the domain of interest.

**Definition 5.4** Let $D = \langle \mathcal{L}_D, \mathcal{R}, \mathcal{AB}, \Gamma \rangle$ be a domain description and $\Delta \subseteq \mathcal{AB}$. A model $I = \langle h, \phi, \eta, \varepsilon_q, \varepsilon_f \rangle$ of $D$ is $\Delta$-relativised iff

1. for each $\delta \in \mathcal{AB}$, $I \models \delta$ iff $\delta \in \Delta$; and

2. $Occ^I = OA_D \cup DAS(\Delta)$, where:

   (a) $OA_D = \{ (\phi(\tau_1), \alpha, \phi(\tau_2)) \in \mathbb{T} \times \mathcal{A}_\nvDash \times \mathbb{T} \mid \underset{\pm}{\leqq} \models [\eta_\nvDash, \tau_\nvDash]\alpha \}$, and

   (b) $DAS(\Delta) = \{ (t, da_{\neg l}, t^+) \in \mathbb{T} \times \mathcal{DA} \times \mathbb{T} \mid [\approx]\mathbb{FA}_\lessdot \notin \underset{\pm}{\geqq}$ and there do not exist any action
   $\alpha \in \mathcal{A}_0$ and $s \in \mathbb{T}$ such that $r = \dfrac{\Phi, [\tau_1, \tau_2]\alpha, [\tau_1, \tau_2]AQ_{\neg l}}{\neg[\tau_2]l \land \neg[\tau_1]FA_l} \in \mathcal{R}$ and $I \models$
   $prem(r)[\tau_1/s, \tau_2/t^+] \}$.

$\Delta$-relativised models are one of the central notions of our framework. Essentially, assumptions underpin our machinery to conjecture information based on common sense knowledge. As such, we will try to accept as many assumption as possible unless there is a good reason not to. Therefore, given a set of assumptions, we will attribute every missing frame assumption to a possible change in the domain the agent is reasoning about which is caused by either a known action/event or some unknown action, called dummy actions in this paper.

The following observation is immediate from condition (1.) in the above definition.

**Observation 3** Let a domain description $D = \langle \mathcal{L}_D, \mathcal{R}, \mathcal{AB}, \Gamma \rangle$ and a set of assumptions $\Delta \subseteq \mathcal{AB}$ be given. If the model $I$ of $D$ is $\Delta$-relativised then $\Delta_{QF}^I = I(\Delta, t)$ for every $t \in \mathbb{T}$.

---

11. The notation $\varphi[v_1/t_1, \ldots, v_n/t_n]$ is standard in logic and meant to be the instantiation of the formula $\varphi$ with the variables $v_1, \ldots, v_n$ being replaced by the terms $t_1, \ldots, t_n$, respectively.

## 5.1 The Frame Problem

First we will address the frame problem in a simple setting *viz.* without qualification assumptions, but we will lift the restrictions later.

**Definition 5.5** Let $D = \langle \mathcal{L}_D, \mathcal{R}, \mathcal{AB}, \Gamma \rangle$ be a domain description. $D$ is a *simple domain description*, or *S-domain*, iff $\mathcal{R}_Q = \emptyset$ and $AQ$ does not occur anywhere in $\mathcal{R}$ or $\Gamma$.

**Definition 5.6** Let $D = \langle \mathcal{L}_D, \mathcal{R}, \mathcal{AB}, \Gamma \rangle$ a domain description. An interpretation $I = \langle h, \phi, \eta, \varepsilon_q, \varepsilon_f \rangle$ is a *simple model*, or *S-model*, of $D$ iff

1. $I$ is a model of $D$; and

2. $\varepsilon_q(t, AQ_l, u) = \textbf{true}$ for every $(t, AQ_l, u) \in \mathbb{T} \times \mathbb{AQ}_{\mathcal{F}^*} \times \mathbb{T}$.

This effectively isolates the frame problem from the qualification problem. Note also that if $I$ is an S-model then $\Delta_{AQ}^I = \mathcal{AB}_{AQ}$. A *coherent S-model* is an S-model which is coherent.

**Example 4** (*continued.*) The following is part of one of the coherent models of $D_{\text{YSP}}$:

$\{[0, 1]load, \neg[0]loaded, [1]loaded, [0]alive, [1]alive,$
$[1, 2]wait, [1, 2]da_{\neg loaded}, \neg[2]loaded, [2]alive,$
$[2, 3]shoot, \neg[3]loaded, [3]alive\},$

which corresponds to one of the anomalous models of this scenario (the one pointed out by Hanks and McDermott).

But it is not desirable to admit the occurrence of an event when there is no evidence for it. Thus we need to minimise the set of action occurrences in a given action theory.

**Definition 5.7** Let $D$ be an S-domain. A coherent S-model $I$ of $D$ is a *prioritised minimal model* (or simply *PMM*) of $D$ iff there does not exist any coherent S-model $I'$ of $D$ such that $Occ^{I'} \subset Occ^I$.

Note that the above model-theoretic minimisation policy is not based on the frame assumptions. This solution to the frame problem is thus amenable to well-known techniques such as circumscription[12], but we believe an argumentation-theoretic approach is not only more direct but also has wider applicability. In order to provide the connection between the above (model-theoretic) minimisation policy and the (argumentation-theoretic) notion of plausible sets of assumptions we need to maximise the set of assumptions satisfiable in a PMM.

**Definition 5.8** Let $D$ be an S-domain. A PMM $I$ of $D$ is a *canonical prioritised minimal model* (or simply *CPMM*) of $D$ iff there does not exist any PMM $I'$ of $D$ such that $FA^I \subset FA^{I'}$.

We now want to see how the account of plausible sets of assumptions connects to this account of minimality.

---

12. in combination with the introduction of occurrences of dummy actions.

**Lemma 1** *Let $D = \langle \mathcal{L}_D, \mathcal{R}, \mathcal{AB}, \Gamma \rangle$ be an S-domain. If $I$ is a CPMM of $D$ then for each assumption $\delta \in \mathcal{AB}$: $\delta \notin \Delta_{QF}^I$ iff $\delta$ is rejected by $\Delta_{QF}^I$.*

*Proof.* ($\Rightarrow$) Suppose by way of contradiction that $\delta$ is not rejected by $\Delta_{QF}^I$, then $\Delta_{QF}^I \cup \{\delta\}$ is $\mathcal{R}$-consistent, i.e. $\Gamma \cup \Delta_{QF}^I \cup \{\delta\} \nvdash_{\mathcal{R}} \textbf{false}$. Since $D$ is an S-domain, $\mathcal{AB}_{AQ} \subseteq \Delta_{QF}^I$. Assume that $\delta = [\tau] FA_l$ for some $\tau \in \mathcal{TE}$ and $l \in \mathcal{F}^*$, we can construct an interpretation $I'$ in such a way that $I'$ interprets everything except $FA$ the same as $I$ and $FA^{I'} = FA^I \cup \{(\phi(\tau), FA_l)\}$.

Since $I$ is a PMM of $D$, from the above construction, $I'$ is also PMM of $D$. But $FA^I \subset FA^{I'}$. Contradiction.

($\Leftarrow$) Obvious.                                                                                   $\square$

**Lemma 2** *Let $D = \langle \mathcal{L}_D, \mathcal{R}, \mathcal{AB}, \Gamma \rangle$ be an S-domain and $I$ a CPMM of $D$. If there are $t \in \mathbb{T}$ and $l \in \mathcal{F}^*$ such that $[t] FA_l \in Lr(\Delta_{QF}^I)$ then $I \models [t, t^+] da_{\neg l}$.*

*Proof.* Let $\delta$ denote the assumption $[t] FA_l$. First we observe that $\delta \in Lr(\Delta_{QF}^I)$ implies $\delta \notin \Delta_{QF}^I$ since $\delta$ is rejected by $\Delta_{QF}^I$ and $\Delta_{QF}^I$ does not attack itself. This in turn implies that $I \models \{[t]l, \neg[t^+]l\}$ as the denotation of $FA$ in $I$ is required to be maximal since $I$ is a CPMM of $D$. From the condition that $I$ is coherent, either (i) there are $\alpha \in \mathcal{A}_0$ and $s \in \mathbb{T}$ such that

$$ r = \frac{\Phi, [\tau_1, \tau_2]\alpha, [\tau_1, \tau_2]AQ_{\neg l}}{\neg[\tau_2]l \wedge \neg[\tau_1]FA_l} \in \mathcal{R} $$

and $I \models prem(r)[\tau_1/s, \tau_2/t^+]$, or (ii) $I \models [t, t^+] da_{\neg l}$. Condition (i) guarantees that $\delta$ is attacked by $\Delta_{QF}^I$ and thus $\delta$ can not be a member of $Lr(\Delta_{QF}^I)$. Therefore, (ii) must be the case.                                                                                   $\square$

The converse of Lemma 2 does not hold. There are cases in which $I \models [t, t^+] da_{\neg l}$ and the assumption $[t] FA_l$ is attacked by $\Delta_{QF}^I$ as a basic action occurs that changes the fluent literal $l$.

**Theorem 1** *Let $D = \langle \mathcal{L}_D, \mathcal{R}, \mathcal{AB}, \Gamma \rangle$ be an S-domain. If $I$ is a CPMM of $D$ then $\Delta_{QF}^I$ is plausible.*

We now prove that not only can we derive a plausible set of assumptions from a given CPMM but we can also construct CPMMs from a plausible set of assumptions of a given S-domain.

The set of $\Delta$-relativised models of an S-domain $D$ is denoted as $Mod_\Delta^S(D)$.

**Observation 4** *Let $D$ be an S-domain and $\Delta$ a set of assumptions of $D$. For each $I \in Mod_\Delta^S(D)$, $\Delta = \Delta_{QF}^I$.*

*Proof.* From the construction of $\Delta$-relativised models:

For each $I \in Mod_\Delta^S(D)$, $\delta = [\tau] FA_l \in \Delta$ iff $I \models [\tau] FA_l$ iff $\delta \in \Delta_{FA}^I$. (More precisely, we have the assumption $[\phi(\tau)] FA_l$ is in $\Delta_{FA}^I$, where $\phi$ is the valuation defined in $I$. But relative to $I$, it is identical to $\delta$.)

Therefore, $\Delta = \Delta_{QF}^I$.                                                                                   $\square$

**Theorem 2** *Let $D = \langle \mathcal{L}_D, \mathcal{R}, \mathcal{AB}, \Gamma \rangle$ be an S-domain and $\Delta \subseteq \mathcal{AB}$. $\Delta$ is plausible wrt $D$ iff $Mod^S_\Delta(D) \neq \emptyset$ and for each $I \in Mod^S_\Delta(D)$, $I$ is a CPMM of $D$.*

**Theorem 3** *Let $D = \langle \mathcal{L}_D, \mathcal{R}, \mathcal{AB}, \Gamma \rangle$ be an S-domain. Furthermore, suppose that $CPMM(D)$ is the set of CPMMs of $D$ and $Plaus(D)$ is the set of plausible sets of assumptions of $D$, then $CPMM(D) = \bigcup_{\Delta \in Plaus(D)} Mod^S_\Delta(D)$.*

### 5.1.1 Discussion:

So, how does our account of the frame problem relate to the existing approaches to the frame problem? While there has been a long line of development behind monotonic approaches to the frame problem starting with Haas's (1987) and Schubert's (1990) early attempts and resulting in Reiter's (1991) monotonic solution to the frame problem together with other solutions proposed by others such as Castilho et al. (1999) and Zhang and Foo (2002) or Thielscher's (1999) Fluent Calculus-based monotonic solution to the frame problem, with the notable exception of Thielscher's (2001) attempt to address the qualification problem, few have tried to tackle the qualification within the framework they use to address the frame problem.

On the other hand, in the action languages (Gelfond & Lifschitz, 1998) and related approaches such as those proposed by McCain and Turner (1995, 1997), Giunchiglia, Kartha, and Lifschitz (1997), and Giunchiglia and Lifschitz (1998), state transition systems are employed as the underlying computation machinery which essentially provides the reasoner with all possible complete states of the world. Furthermore, as domain decsriptions can be uniquely translated to state transition systems, the reasoner could safely derive the successor state(s) based on the current together with the transition function.

## 5.2 Solving the Qualification Problem (in the Presence of the Frame Problem)

The results reported in the previous section are established in a simple setting. If we add the following observation to the theory in example 4: $[3]alive$, i.e. after the *shoot* action, the victim is still alive, then like most existing formalisms, the above account of plausibility would come up with a contradiction. In fact, it would be more reasonable that such a failure is explained as an occurrence of some qualification. In this section, we remove certain restrictions on the qualifications of actions in order to achieve a more general framework.

There are some subtleties in the way action theories are represented in our proposed assumption-based framework. Note first that there is a potential difficulty if frame assumptions and qualification assumptions are treated equally, which can be illustrated by a version of the YSP. Consider the following action description:

$$\{ \frac{[\tau]alive, [\tau]FA_{alive}}{[\tau^+]alive}, \frac{[\tau]loaded, [\tau, v]shoot, [\tau, v]AQ_{\neg alive}}{\neg[v]alive \wedge \neg[\tau]FA_{alive}} \} \subseteq \mathcal{R},$$

$$\{[0]loaded, [0]alive, [0, 1]shoot\} \subseteq \Gamma.$$

From this, we have (at least) two stable sets of assumptions: one contains the frame assumption $[0]FA_{alive}$ which rejects the qualification assumption $[0, 1]AQ_{\neg alive}$ and another contains $[0, 1]AQ_{\neg alive}$ which attacks $[0]FA_{alive}$. Only the latter is intuitive in this case but we do not have any explicit criterion to prefer one over another. The following assumption asserts that in our solution to the frame problem in the presence of the qualification problem, an action is presumed to bring about its effects unless there is an explicit justification for its disqualification.

**Assumption 2** *When there is a direct conflict between a frame assumption and a qualification assumption (over a fluent literal), the qualification assumption takes precedence.*

Given the presence of several kinds of assumptions, i.e. frame and qualification, we will adopt the following convention: we will write $Lr_P(\Delta)$ instead of $(Lr(\Delta))_P$ for $P \in \{FA, AQ\}$. Since we no longer exclude qualification assumptions from our assumption-based domain descriptions, we will simply refer to assumption-based domain descriptions as *Q-domains*.

**Definition 5.9** Let $D = \langle \mathcal{L}_D, \mathcal{R}, \mathcal{AB}, \Gamma \rangle$ be a Q-domain. A presumable set of assumptions $\Delta \subseteq \mathcal{AB}$ is *semi-Q-plausible* wrt $D$ iff $Lr_{FA}(\Delta)$ is minimal (with respect to set inclusion).

**Definition 5.10** Let $D = \langle \mathcal{L}_D, \mathcal{R}, \mathcal{AB}, \Gamma \rangle$ be a Q-domain. A set of assumptions $\Delta \subseteq \mathcal{AB}$ is *Q-plausible* wrt $D$ iff

1. $\Delta$ is semi-Q-plausible wrt $D$,

2. $\Delta_{AQ}$ is maximal, i.e. there does not exist any $\Delta' \subseteq \mathcal{AB}$ such that $\Delta'$ is semi-Q-plausible (wrt $D$) and $\Delta_{AQ} \subset \Delta'_{AQ}$, and

3. $\Delta_{FA}$ is maximal relative to the above two conditions, i.e. there does not exist any $\Delta' \subseteq \mathcal{AB}$ such that $\Delta'$ satisfies the above two conditions and $\Delta_{FA} \subset \Delta'_{FA}$.

We will now refer to models of a Q-domain as *Q-models*. A *coherent Q-model* is a Q-model which is coherent. We minimise the set of action occurrences in coherent Q-models of a given action theory.

**Definition 5.11** Let $D$ be a Q-domain. A coherent Q-model $I$ of $D$ is a *prioritised minimal Q-model* (or simply *PMQM*) of $D$ iff there does not exist any coherent Q-model $I'$ of $D$ such that $Occ^{I'} \subset Occ^{I}$.

**Definition 5.12** Let $D$ be an S-domain. A PMQM $I$ of $D$ is a *canonical prioritised minimal Q-model* (or simply *CPMQM*) of $D$ iff

1. there does not exist any PMQM $I'$ of $D$ such that $AQ^I \subset AQ^{I'}$, and

2. there does not exist any PMM $I'$ of $D$ such that $FA^I \subset FA^{I'}$.

Now we can proceed to obtaining the main results for CPMQMs regarding Q-plausible sets of assumptions which are similar to those for CPMMs regarding plausible sets of assumptions. The following lemma, which is a straightforward extension of Lemma 1 and Lemma 2 proved in the previous section, is introduced to assist in the proof of Theorem 4 .

**Lemma 3** *Let $D = \langle \mathcal{L}_D, \mathcal{R}, \mathcal{AB}, \Gamma \rangle$ be an Q-domain and $I$ a CPMM of $D$,*

1. For each assumption $\delta \in \mathcal{AB}$: $\delta \notin \Delta^I_{QF}$ iff $\delta$ is rejected by $\Delta^I_{QF}$.

2. If $\delta = [\tau]FA_l \in Lr(\Delta^I_{QF})$ then $I \models [\tau, \tau^+]da_{\neg l}$.

**Theorem 4** *Let $D$ be a Q-domain. If $I$ is a CPMQM of $D$ then $\Delta^I_{QF}$ is Q-plausible wrt $D$.*

Similar to the previous section, we now prove that not only can we derive a plausible set of assumptions from a given CPMQM but we can also construct CPMQMs from a plausible set of assumptions of a given domain description. The set of $\Delta$-relativised models of a Q-domain $D$ is denoted as $Mod^Q_\Delta(D)$.

The following observation is obvious:

**Observation 5** *Let $D$ be a Q-domain and $\Delta$ a set of assumptions of $D$. For each $I \in Mod^Q_\Delta(D)$, $\Delta = \Delta^I_{QF}$.*

**Theorem 5** *Let $D = \langle \mathcal{L}_D, \mathcal{R}, \mathcal{AB}, \Gamma \rangle$ be a Q-domain and $\Delta \subseteq \mathcal{AB}$. $\Delta$ is Q-plausible wrt $D$ iff $Mod^Q_\Delta(D) \neq \emptyset$ and for each $I \in Mod^Q_\Delta(D)$, $I$ is a CPMQM of $D$.*

**Theorem 6** *Let $D$ be a Q-domain. Furthermore, suppose that $CPMQM(D)$ is the set of CPMQMs of $D$ and $Plaus^Q(D)$ is the set of Q-plausible sets of assumptions of $D$, then $CPMQM(D) = \bigcup_{\Delta \in Plaus^Q(D)} Mod^Q_\Delta(D)$.*

Q-plausible sets of assumptions allow one to overcome scenarios in which expectation failures (or, qualification surprises) arise, e.g. shooting a turkey with a loaded gun and observing that the turkey is still alive. When such surprises arise, the reasoner knows who's to blame: qualification assumptions. She can then accordingly remove the "guilty" assumptions. Just as the anomalous models forming obstacle to early approaches to the frame problem, similar anomalous models can also arise in solutions to the qualification problem. This important issue related to the qualification problem has been thoroughly discussed by Thielscher (2001) and a solution was presented within the framework of the Fluent Calculus. To give the reader a flavour of this problem within our framework, we invite the reader to consider the following classical example:

**Example 5** Consider the problem of starting a car whose tail pipe could possibly be blocked by a potato, formalised in our formalism as follows.

1. The set of inference rules $\mathcal{R}$ is:[13]

$$\frac{[\tau]BlockedTP}{\neg[\tau,v]AQ_{GetStarted}},$$

$$\frac{[\tau]HasPotato, [\tau,v]InsertPotato, [\tau,v]AQ_{BlockedTP}}{[v]BlockedTP \wedge \neg[\tau]FA_{\neg BlockedTP}},$$

$$\frac{[\tau]HasKey, [\tau,v]TurnOnIgnition, [\tau,v]AQ_{GetStarted}}{[v]GetStarted \wedge \neg[\tau]FA_{\neg GetStarted}}.$$

---

13. Of course we also have the frame-based inference rules for $HasKey$, $BlockedTP$, etc. but we omit them from this representation for sake of readability.

2. The theory $\Gamma$ is: $\{[0]HasPotato, [0]HasKey, \neg[0]BlockedTP, \neg[0]GetStarted, [0,1]InsertPotato, [1,2]TurnOnIgnition\}$.

Of course, we have designed this example to try to avoid any possible troubles with the frame problem. We have to consider two conflicting qualification assumptions in this case which are $[0,1]AQ_{BlockedTP}$ and $[1,2]AQ_{GetStarted}$. Given the above action theory, any Q-plausible set of assumptions would contain exactly one of them. Thus, there will be at least two extensions, one in which the car can not get started since the tailpipe is blocked and it's no longer consistent to assume $[1,2]AQ_{GetStarted}$. The other extension disqualifies the action of inserting the potato into the tailpipe and thus the action of starting the car becomes successful. Only the former is intuitive in this case and the account of Q-plausible sets of assumptions fails to deliver this desired solution.

However, in exactly the same way the problem of anomalous models arising in the solution to the frame problem is tackled, the above problem is easily addressed within our framework. The key insight is of course also underlined by the notion of causality: $[1]BlockTP$ was caused (by the action $[0,1]InsertPotato$) which in turn allows $\neg[1,2]AQ_{GetStarted}$ to be derived. On the other hand, there was no cause that allows $\neg[0,1]AQ_{BlockTP}$ to be derived. How is the above insight realised in our framework? The answer turns out to be rather simple: Just as the distinction between leniently rejected frame assumptions and non-leniently rejected frame assumptions allows us to distinguish between normal changes (i.e. action-triggered) and anomalous changes, a distinction between leniently rejected qualification assumptions and non-leniently rejected qualification assumptions will allow us to distinguish between normal disqualifications (i.e. those underlined by a cause) and anomalous disqualifications. Thus, facing a collection of Q-plausible sets of assumptions, a reasoner simply selects the set of assumptions that contains the smallest (with respect to set inclusion) set of leniently rejected qualification assumptions.

The ability to introduce different argumentation-theoretic semantics for assumption-based frameworks is arguably the biggest advantage of our approach. The following example illustrates this critical point:

**Example 6** We modify an example presented by Lin and Shoham (1995) which is in turn a modification of Kautz's (1986) Stolen Car Problem. A spy possessed a microfilm of a top secret evidence which an organisation, $A$, tried to steal. For some reason, another organisation, $B$, wanted to murder this spy. The microfilm was in the safe at the spy's home at time 0. The spy was not at home at time 0. $A$ tried to *steal* the evidence between the time points 0 and 1. The spy might *return* home at any time between 0 and 1. $B$ tried to *murder* the spy at any time between 0 and 1. *return* cancels the effects of *steal*, *murder* cancels the effects of *return* and *steal* cancels the effects of *murder*. Any of the three actions *steal*, *return*, and *murder* takes only one time step. The domain is formalised as follows: $\Gamma = \{\neg[0]EvStolen, [0]Alive, \neg[0]AtHome\}$; and $\mathcal{R}$ contains

$$\{\frac{[\tau,\upsilon]steal, [\tau,\upsilon]AQ_{EvStolen}}{[\upsilon]EvStolen \wedge \neg[\tau]FA_{\neg EvStolen}}, \frac{[\tau,\upsilon]return, [\tau,\upsilon]AQ_{AtHome}}{[\upsilon]AtHome \wedge \neg[\tau]FA_{\neg AtHome}}, \frac{[\tau,\upsilon]murder, [\tau,\upsilon]AQ_{\neg Alive}}{\neg[\upsilon]Alive \wedge \neg[\tau]FA_{Alive}},$$

$$\frac{[\tau]AtHome \wedge \tau < \upsilon}{\neg[\tau,\upsilon]AQ_{EvStolen}}, \frac{\neg[\tau]Alive \wedge \tau < \upsilon}{\neg[\tau,\upsilon]AQ_{AtHome}}, \frac{[\tau]EvStolen \wedge \tau < \upsilon}{\neg[\tau,\upsilon]AQ_{\neg Alive}}\}.$$

Given the above formalisation of the problem, traditional accounts of non-monotonic reasoning (i.e. Default Logic, circumscription, Autoepistemic Logic, etc.) can not provide one with a solution since these formalisms can not produce any extension under their standard semantics. However, the argumentation-theoretic approach does gives several semantics for this problem including preferability semantics. Note, however, that all admissible sets of assumptions and preferred sets of assumptions for this domain contain **none** of the assumptions: $[0,1]AQ_{EvStolen}$, $[0,1]AQ_{AtHome}$, and $[0,1]AQ_{\neg Alive}$. This essentially means that, under the admissibility and the preferability semantics, the reasoner could only infer that none of the above actions would succeed On the other hand, our proposed plausible semantics gives an alternative solution for this problem in which each consistent set of assumptions can contain at most one of the following three assumptions: $[0,1]AQ_{EvStolen}$, $[0,1]AQ_{AtHome}$, and $[0,1]AQ_{\neg Alive}$. Moreover, each plausible set of assumptions must contain exactly one of them. Among the other two assumptions which do not belong to a plausible set of assumptions, one is attacked and the other is leniently rejected.

We believe that the above examples have underlined the major advantages of our approach in which a reasoner is aware of the (defeasible) assumptions used in her reasoning as well as being able to explicitly reason about these assumptions. The flexibility of allowing a reasoner to introduce different argumentation-theoretic semantics for assumption-based framework by simply varying the notion of acceptability of sets of assumptions is certainly another advantage in favour of our approach.

## 6. More Complex Dynamic Domains and Indirect Effects

So far we haven't taken into consideration the issues of concurrent actions and indirect effects. To ensure that the formalisation introduced is expressive enough to deal with complex domains, we will show how these issues are coped with by our approach. Firstly, we motivate our formalisation with an informal discussion.

### 6.1 Concurrent and Non-Deterministic Events

Given our temporal representation, formulating concurrent events is not a difficult issue in our framework. However, there are some subtleties which need to be carefully considered. Firstly, the use of assumptions. As presented earlier in this paper, qualification assumptions are fluent oriented, i.e. we qualify over the effects of action rather than over the action itself. Whilst this manifests the capability of formulating actions with multiple effects, and thus each effect should be qualified independently, it may fail to formalise concurrent events with the same effects. For example, both actions *hit* the vase with a hammer and *shoot* it with a loaded gun bring about the effect that the vase is *broken*. In other words, it is essential that qualification assumptions be dependent on the actions that bring about the effect under consideration. Thus, given $n$ actions and $m$ fluents whose values can be changed by those actions, we have to potentially introduce $2 \times m \times n$ qualifications assumptions.[14] Therefore, instead of subscripting the assumption symbols $AQ$ with the fluent literals from $\mathcal{F}^*$, we extend the set of subscripts of $AQ$, denoted as $\mathcal{AF}$, to contain both the action and the corresponding fluent literals.[15] For example, given the above two actions *hit* and *shoot* and an additional action *repair* whose effect is change a broken vase to being non-broken, *viz.* ¬*broken*, we will need to introduce the following qualification assumptions: $AQ_{Hit\text{-}Broken}, AQ_{Shoot\text{-}Broken}$ and $AQ_{Repair\text{-}\neg Broken}$. Therefore, syntactically we also extend the set $\mathcal{AB}_{AQ} = \{[\tau, v]AQ_\varphi \mid \tau, v \in \mathcal{TE} \text{ and } \varphi \in \mathcal{AF}\}$ and semantically the function $\varepsilon_q : \mathbb{T} \times \mathbb{AQ}_{\mathcal{AF}} \times \mathbb{T} \to \{\textbf{true}, \textbf{false}\}$. In the definition of an interpretation $I = \langle h, \phi, \eta, \varepsilon_q, \varepsilon_f \rangle$, we also have $I([\tau, v]AQ_\varphi, t) = \varepsilon_q(\phi(\tau), AQ_\varphi, \phi(v))$, where $\tau, v \in \mathcal{TE}$ and $\varphi \in \mathcal{AF}$.

---

14. In fact, as we will see later, there are potentially $2 \times m \times (n+1)$ qualification assumptions in this case since there is one special event corresponding to all (natural) events that bring about the indirect effects.

Though this does increase the complexity of the framework, it is the price we have to pay for the expressiveness of the resulting system. To the best of our knowledge, none of the existing formalisms possesses such an expressiveness.

How about non-deterministic actions? The solution turns out to be quite simple: we can treat action non-determinism as a special kind of action qualification. For example, to formulate the Russian shooting scenario (Sandewall, 1994) in which a gun non-deterministically gets loaded or not after spinning its revolver, the following action descriptions can be asserted:

$$adr_1 = \frac{[\tau, v]spin, [\tau, v]AQ_{Spin\text{-}Loaded}}{[v]loaded \wedge \neg[\tau]FA_{\neg loaded}} \text{ and } adr_2 = \frac{[\tau, v]spin, [\tau, v]AQ_{Spin\text{-}\neg Loaded}}{\neg[v]loaded \wedge \neg[\tau]FA_{loaded}}$$

together with the following two qualification rules:

$$qr_1 = \frac{[v]loaded}{\neg[\tau, v]AQ_{Spin\text{-}\neg Loaded}} \text{ and } qr_2 = \frac{\neg[v]loaded}{\neg[\tau, v]AQ_{Spin\text{-}Loaded}}.$$

The above guarantee that either ¬$[v]loaded$ or $[v]loaded$ will follow from $[\tau, v]spin$ (remark that the set of qualification assumptions $\Delta_{AQ}$ is maximised), but not both. As a consequence, two possible extensions will arise given the above non-deterministic action.

Dealing with non-determinism could be more complicated when the information used to encode the action description is disjunctive. As we only restrict the conclusion of an action description rule to contain only fluent literals, such disjunctive effect is not straightforwardly dealt with in our framework. However, note that the restriction is similar to that imposed on action language $\mathcal{A}$ (Gelfond & Lifschitz, 1998). An extension to action descriptions which is similar to the way the action language $\mathcal{A}$ is extended into the action language $\mathcal{C}$ can be done to allow a complex expression in the conclusion of action description rule. However, for the sake of presentation, we choose to use a simpler language to introduce our framework to the reader.

## 6.2 Formalisation

From the analysis in the introductory section regarding indirect effects, we observe that the ordering in which domain constraints are applied plays an essential role in a technically sound framework. Moreover, it's no longer guaranteed that domain constraints would be strictly satisfied at every time point.[16] To help the reader better understand this technical subtlety, it's useful to think that changes are attributable to events. However, events are further divided into two categories: external and internal. Basically, *external events* correspond to the direct effects of actions performed by some agents (including the reasoner.) On the other hand, *internal events* correspond to the indirect effects when certain conditions about the world are met and can be attributed to Nature.[17]

---

15. The reader is referred to section 3 for the general formalisation.
16. Of course, it's not necessary that the state of the world at every time point is observable to a reasoner. However, it is important that she be aware of such states and able to reason about them.

Like external events, internal events also happen in a certain order. Although it is not straightforward to observe this order[18], it is important that a reasoner be able to reason about them. Hence we propose to add one more dimension into the set of assumptions of a given domain description. Since these assumptions play essentially the same role as that of qualification assumptions,[19] we can subsume them under the set of qualification assumptions $AQ_{\mathcal{AF}}$. As they are not associated with any specific action, we can ignore the initial of the action name from the subscripts of $AQ$. For example we will write $AQ_{broken}$ in case we want to qualify over the fluent *broken* as an indirect effect, i.e. independently of any action. Moreover, to avoid the axiomatiser from the confusion of determining the time span taken by indirect effects, we will assume that it is atomic. It means all indirect effects always take place between one time point and the next one. This does not mean that all indirect effects have the same real-time duration, it simply means that relative to the given time structure **T** they take place between two consecutive time points. This allows us to avoid the granularity problem as it is irrelevant from the viewpoint of the problems we are trying to solve.

### 6.2.1 A Representation Issue and Some Notations:

One important remark is in place here. Until now, we have used integers and the corresponding expressions to denote time points. There are two reasons for this pratice: (i) It significantly simplifies the presentation, and (ii) Without the complication of the ramification problem, all time points which are reasoned about are also observable to the reasoner. However, as will be discussed thoroughly in the next section, in the presence of ramifications a time point at which some change takes place might not be observable to the reasoner as such a change could be one of the indirect effects after the execution of some action. Such representation issues make the integer-based time structure employed so far in this paper inappropriate. We do need a richer representation of time structure. Following Sandewall's (1994) basic formulation, the only basic constant to be included in the representation of the time structure **T** is denoted by the symbol $\Theta$ and is referred to as the *origin* of **T**. The standard algebraic operations $+$ and $-$ are still employed to obtain time expressions and bear their usual meanings. The relation symbol $<$ will also be used to compare time expressions. However, we can no longer allow expressions such as $\Theta + 1$ for integers are no longer elements of the time structure.

Given a time expression $\tau$:

---

17. We avoid the use of the terms exogenous/endogenous events to label these two categories because throughout the literature of reasoning about action, exogenous events are used to refer to actions that carried out by agents that are different from the reasoner or outside events whose occurrences are beyond the reasoner's control.

18. Unless indirect effects are somehow delayed and become observable to the reasoner, they usually take place immediately after direct effects.

19. The only difference is that these assumptions are qualified over the indirect effects which can be considered to be the effects of the actions of Nature.

- We will continue to denote the next time point of $\tau$ by $\tau^+$.

- We define $\tau^{1+}$ to be $\tau^+$ and, let $n > 1$, $\tau^{n+}$ to be $(\tau^{(n-1)+})^+$.

Furthermore, as we will assume throughout that an indirect effect takes place between two consecutive time points, we also simplify the presentation by using the following syntactical sugar: instead of writing $[\tau, \tau^+]AQ_l$ (for some $\tau \in \mathcal{TE}$ and $l \in \mathcal{F}^*$), we will write $[\tau]AQ_l$. This also provides a simple way to distinguish between qualification assumptions for ramificational effects and those caused directly by an action or event.

How are domain descriptions affected from this augmentation? The only change is for the set $\mathcal{R}_A$ of action descriptions which consists of rules either of the form

$$\frac{\Phi, [\tau, v]\alpha, [\tau, v]AQ_{\alpha \leftarrow l}}{[v]l \wedge \neg[\tau]FA_{\neg l}}$$

or of the form

$$\frac{\Phi, [\tau]AQ_l}{[\tau^+]l \wedge \neg[\tau]FA_{\neg l}}.$$

For convenience, we will refer to the set of inference rules having the latter form as $\mathcal{R}_I$, i.e. $\mathcal{R}_I = \{r \in \mathcal{R}_A \mid r = \dfrac{\Phi, [\tau]AQ_l}{[\tau^+]l \wedge \neg[\tau]FA_{\neg l}}$ for some fluent literal $l \in \mathcal{F}^*\}$. Now, regarding the ramification problem, the basic idea is that the state obtained by updating the previous state may not necessarily be stable due to the presence of indirect effects. By representing indirect effects as causal rules (using inference rules in our framework) we can reason about which causal rules have fired and (relatively) when. Moreover, since these causal rules may not take the same amount of time to fire, we should be able to reason about different possible orders in which they fire. This will be achieved by a distinction between stable and unstable states.

**Definition 6.1** A time-point expression $\theta \in \mathcal{TE}$ is *stable* wrt a given domain description $D = \langle \mathcal{L}_D, \mathcal{R}, \mathcal{AB}, \Gamma \rangle$ and $\Delta \subseteq \mathcal{AB}$ iff there does not exist any fluent literal $l \in \mathcal{F}^*$ such that

1. $\dfrac{\Phi, [\tau]AQ_l}{[\tau^+]l \wedge \neg[\tau]FA_{\neg l}} \in \mathcal{R}_I$ and $\Gamma \cup \Delta \vdash_{\mathcal{R}} \Phi, [\theta]AQ_l$, and

2. $\Gamma \cup \Delta \nvdash_{\mathcal{R}} [\theta]l$.

**Definition 6.2** Let $D = \langle \mathcal{L}_D, \mathcal{R}, \mathcal{AB}, \Gamma \rangle$ be a domain description. A set of assumptions $\Delta$ is said to be *ramification-compliant* iff

1. there does not exist any $\delta \in \mathcal{AB}_{AQ}$ such that $Th_{\mathcal{R}}(\Delta \cup \{\delta\}) \neq Th_{\mathcal{R}}(\Delta) \cup \{\delta\}$.

2. there does not exist any unstable time-point expression $\tau \in \mathcal{TE}$ such that $[\tau]FA_l \in \Delta$ for every $l \in \mathcal{F}^*$.

We note that by using unstable time points, we have conceptually isolated the ramification problem from the task of reasoning about the (explicit) actions.

**Definition 6.3** Let $D = \langle \mathcal{L}_D, \mathcal{R}, \mathcal{AB}, \Gamma \rangle$ be a domain description. A set of assumptions $\Delta$ is *generally plausible for action domains*, or simply *AD-plausible*, iff

- $\Delta$ is ramification-compliant, and

- $\Delta$ is Q-plausible relative to the above condition.

Remarks:

1. From the above definition, it can be seen that a solution for the ramification is (in a sense) independent from the frame and the qualification problems regarding action occurrences.

2. Given the above temporal ontology, it is worth noting that the traditional notion of state constraints may no longer hold in this representation regarding time points. More precisely, only states associated with stable time points are subject to these constraints.

## 6.3 Connection to Causation-Based Formalisms

The question is, of course, whether the above computational mechanism gives satisfactory conclusions for problems in reasoning about action. While many formalisms have been proposed, a general criterion for reasoning about action formalisms still seems to be missing. The major stream of research towards a solution to the ramification problem is based on the notion of causality (e.g., Lin, 1995; McCain & Turner, 1995; Thielscher, 1997). As has been discussed above, none of these approaches can deal with domains in which (potentially) infinite sequences of indirect effects are present. Most of these approaches (including the above three references), however, produce a successor state after the execution of an action to capture changes that have taken place either as direct or as indirect effects of that action. In this sub-section, we show how our formalism captures the notion of successor states in the absence of the infinite sequences of indirect effects.

**Definition 6.4** A domain description $D = \langle \mathcal{L}_D, \mathcal{R}, \mathcal{AB}, \Gamma \rangle$ is *non-stratified* iff there exist two sets $\Phi \subseteq \Lambda$ and $\{l_1, \ldots, l_n\} \subseteq \mathcal{F}^*$ such that

1. for each $1 \le i \le n$, $\Phi \cup \{[\tau]l_i\}$ is $\mathcal{R}$-consistent, for every $\tau \in \mathcal{TE}$; and

2. for each $1 \le i \le n$, if $k = (i \bmod n) + 1$, then there exists a set $\Psi^k \subseteq \Lambda$ such that $\Phi \cup \{[\tau]l_i\} \vdash_{\mathcal{R}} \Psi^k$, and

$$\frac{\Psi^k, [\tau]AQ_{l_k}}{[\tau^+]l_k \wedge \neg[\tau]FA_{\neg l_k}} \in \mathcal{R}_I.$$

Of course, a domain description is *stratified* if and only if it is not non-stratified. Given a domain description $D = \langle \mathcal{L}_D, \mathcal{R}, \mathcal{AB}, \Gamma \rangle$, we'll denote $E_D(\Delta) \overset{def}{=} Th_{\mathcal{R}}(\Gamma \cup \Delta)$, the extension of an action theory $D$ according to $\Delta$. We will also write $E_D$ instead of $E_D(\emptyset)$ for brevity.

**Definition 6.5** Let $\sigma = \langle \mathcal{T}, \mathcal{F}, \mathcal{A} \rangle$ be a signature.

1. A set $S \subseteq \mathcal{F}^*$ is an *instantwise state* iff for every $l \in \mathcal{F}^*$, either $l \in S$ or $\neg l \in S$. $IS$ will be used to denote the set of instantwise states of $\Lambda$,

2. Let $\Gamma \subseteq \mathcal{F}^*$ and $\tau \in \mathcal{TE}$, we denote $\overline{\Gamma} \stackrel{def}{=} \{\neg l \mid l \in \Gamma\}$ and $[\tau]\Gamma \stackrel{def}{=} \{[\tau]l \mid l \in \Gamma\}$,

3. Let $D = \langle \mathcal{L}_D, \mathcal{R}, \mathcal{AB}, \Gamma \rangle$ be a domain description. $D$ is *simplistic* iff $\Gamma = \emptyset$.

The motivation behind the introduction of instantwise states is to allow the reasoner to reason about the intermediate states which may not be practically observable to her. That is, when indirect effects (following an occurrence of an action or event) take place, the world might transit through a number of intermediate states before becoming stabilised in a final state in which all domain constraints necessarily hold. The ability to explicitly reason about such (unstable) intermediate states is one of the advantages offered by our approach. For instance, let's consider the scenario in Example 1 which was originally introduced by Thielscher (1997). Thielscher's (1997) approach allows two possible successor states as the results of closing the switch $sw_1$ in the initial state $S = \{\neg sw_1, sw_2, sw_3, \neg relay, \neg light, \neg detect\}$. These are $T_1 = \{sw_1, \neg sw_2, sw_3, relay, \neg light, \neg detect\}$ and $T_2 = \{sw_1, \neg sw_2, sw_3, relay, \neg light, detect\}$. However, to an outside observer who has no idea about the tricky internal mechanism of this circuit, it could be quite difficult to explain why $T_2$ should be one of the possible outcomes of the action of closing the switch $sw_1$: starting from an initial state in which $detect$ and $light$ are both off, after performing an action $toggle(sw_1)$, $light$ remains off but somehow $detect$ becomes on. In other words, Thielscher's (1997) framework fails to render the intermediate (and unstable) state in which $light$ was on, albeit for only a short instant, before it was turned off again. On the other hand, such states are explicitly represented and reasoned about as instantwise states in our framework. As a consequence, domain constraints do not necessarily hold in an instantwise state.

In the following, we abbreviate simplistic and stratified domain descriptions as *SSD*s. Let $\Omega \subseteq \mathcal{R}_I$, we denote:

$$CONS_R(\Omega) \stackrel{def}{=} \{l \in \mathcal{F}^* \mid \frac{[\tau]\Psi, [\tau]AQ_l}{[\tau^+]l \wedge \neg[\tau]FA_{\neg l}} \in \Omega\}.$$

Similarly, let $\Omega \subseteq \mathcal{R}_A$, we denote

$$CONS_A(\Omega) \stackrel{def}{=} \{l \in \mathcal{F}^* \mid \frac{[\tau_1]\Psi, [\tau_1, \tau_2]\alpha, [\tau_1, \tau_2]AQ_{\alpha \cdot l}}{[\tau_2]l \wedge \neg[\tau_1]FA_{\neg l}} \in \Omega\}.$$

The action description $\frac{[\tau_1]\Psi, [\tau_1, \tau_2]\alpha, [\tau_1, \tau_2]AQ_{\alpha \cdot l}}{[\tau_2]l \wedge \neg[\tau_1]FA_{\neg l}} \in \mathcal{R}_A$ is *applicable* in $S$ iff $\Psi \subseteq S$. We use $\Upsilon_\alpha^S$ to denote the set of action descriptions that is applicable (in $S$) regarding the action $\alpha$. $\Omega \subseteq \Upsilon_\alpha^S$ is a *possible application* of $\alpha$ in $S$ iff $CONS_A(\Omega)$ is consistent and there does not exist any $\Omega' \subseteq \Upsilon_\alpha^S$ such that $\Omega \subset \Omega'$ and $CONS_A(\Omega')$ is consistent.

**Definition 6.6** Let $\sigma = \langle \mathcal{T}, \mathcal{F}, \mathcal{A} \rangle$ be a signature and $D = \langle \mathcal{L}_D, \mathcal{R}, \mathcal{AB}, \emptyset \rangle$ a SSD. Suppose $S \in IS$ and $\alpha \in \mathcal{A}$. We formalise the *direct effects* of an action $\alpha$ using a relation $Res_D$: for all $S, S' \in IS$, $(S, \alpha, S') \in Res_D$ iff there is a possible application $\Omega$ of $\alpha$ in $S$ such that:

(i) $CONS_A(\Omega) \subseteq S'$, and

(ii) there does not exist any instantwise state $S''$ such that $S''$ satisfies (i) and $S'' \setminus S \subset S' \setminus S$.

**Definition 6.7** Let $\sigma = \langle \mathcal{T}, \mathcal{F}, \mathcal{A} \rangle$ be a signature and $D = \langle \mathcal{L}_D, \mathcal{R}, \mathcal{AB}, \emptyset \rangle$ a SSD. The *causation relation* according to $D$, denoted as $Causes_D$, is defined as follows: for all $S, S' \in IS$, $Causes_D(S, S')$ iff there exists a non-empty set $\Omega \subseteq \mathcal{R}_I$ of ramification inference rules such that

(i) for each $\dfrac{[\tau]\Psi, [\tau]AQ_l}{[\tau^+]l \wedge \neg[\tau]FA_{\neg l}} \in \Omega$, $\Psi \subseteq S$ and

(ii) $CONS_R(\Omega) \subseteq S'$, and

(iii) there does not exist any instantwise state $S''$ such that $S''$ satisfies (i) and (ii) and $S'' \setminus S \subset S' \setminus S$.

Given a domain description $D$, a state $S \in IS$ is *stable* regarding $D$ iff there does not exist any $S' \in IS$ such that $Causes_D(S, S')$.

**Definition 6.8** Let $\sigma = \langle \mathcal{T}, \mathcal{F}, \mathcal{A} \rangle$ be a signature and $D$ a SSD. Suppose $w \in IS$ and $\alpha \in \mathcal{A}$. The *state transition* from $w$ in $D$ according to $\alpha$, denoted $Trans_D^\alpha(w)$, is the transitive closure of $Causes_D$ regarding the state $\omega_1 \in IS$ satisfying $(w, \alpha, \omega_1) \in Res_D$. Formally, $Trans_D^\alpha(w) = \{\varpi \in IS \mid$ there exists a sequence $\omega_1, \dots, \omega_n \in IS$ such that $(w, \alpha, \omega_1) \in Res_D$ and $\varpi = \omega_n$ and $(\omega_i, \omega_{i+1}) \in Causes_D$ for each $1 \leq i < n$ and $\varpi$ is stable regarding $D\}$.

Let $I$ be an interpretation for $\mathcal{L}_D$ and $t \in \mathbb{T}$,

- we use $[I]^t$ to denote the instantwise state specified by $I$ at time point $t$: $[I]^t \overset{def}{=} \{l \in \mathcal{F}^* \mid I(l, t) = \mathbf{true}\}$. If $[I]^t$ is stable then $t$ is said to be a *stable time point* in $I$.

- we use $N_I$ to denote a function that maps a time point $t \in \mathbb{T}$ to the next stable time point in $I$: $N_I(t) \in \mathbb{T}$ such that $[I]^{N_I(t)}$ is stable and for every $u \in \mathbb{T}$, if $t < u < N_I(t)$ then $[I]^u$ is not stable.

**Theorem 7** Let $\sigma = \langle \mathcal{T}, \mathcal{F}, \mathcal{A} \rangle$ be a signature and $D_0 = \langle \mathcal{L}_D, \mathcal{R}, \mathcal{AB}, \emptyset \rangle$ a SSD and $\alpha \in \mathcal{A}_l$. Suppose that $w \in IS$. Define a domain description $D = \langle \mathcal{L}_D, \mathcal{R}, \mathcal{AB}, \Gamma \rangle$, where $\Gamma = \{[\Theta]\varphi \mid \varphi \in w\} \cup \{[\Theta, \Theta^+]\alpha\}$. A set $\Delta \subseteq \mathcal{AB}$ of assumptions is AD-plausible wrt $D$ iff for each model $M$ of $E_D(\Delta)$, $[M]^{N_M(\Theta)} \in Trans_D^\alpha([M]^\Theta)$, i.e. $[M]^{N_M(\Theta)}$ belongs to the state transition from $[M]^\Theta$ in $D$ according to $\alpha$.

We now re-consider the motivating examples introduced in Section 1.3.

**Example 1** (*continued.*) We re-formulate the action theory for this example in terms of our formalism:

$$\frac{[\tau]sw_1, [\tau]sw_2, [\tau]AQ_{light}}{[\tau^+]light \wedge \neg[\tau]FA_{\neg light}},$$

$$\frac{\neg[\tau]sw_i, [\tau]AQ_{\neg light}}{\neg[\tau^+]light \wedge \neg[\tau]FA_{light}}(i=1,2),$$

$$\frac{[\tau]relay, [\tau]AQ_{\neg sw_2}}{\neg[\tau^+]sw_2 \wedge \neg[\tau]FA_{sw_2}},$$

$$\frac{\neg[\tau]sw_i, [\tau,v]toggle_i}{[v]sw_i \wedge \neg[\tau]FA_{\neg sw_i}}(i=1,2,3),$$

$$\frac{[\tau]\zeta, [\tau]FA_\zeta}{[\tau^+]\zeta}, \text{ where } \zeta \in \mathcal{F}^*.$$

$$\frac{[\tau]sw_1, [\tau]sw_3, [\tau]AQ_{relay}}{[\tau^+]relay \wedge \neg[\tau]FA_{\neg relay}},$$

$$\frac{\neg[\tau]sw_i, [\tau]AQ_{\neg relay}}{\neg[\tau^+]relay \wedge \neg[\tau]FA_{relay}}(i=1,3),$$

$$\frac{[\tau]light, [\tau]AQ_{detect}}{[\tau^+]detect \wedge \neg[\tau]FA_{\neg detect}},$$

$$\frac{[\tau]sw_i, [\tau,v]toggle_i}{\neg[v]sw_i \wedge \neg[\tau]FA_{sw_i}}(i=1,2,3),$$

The theory is described as follows:

$$\Gamma = \{\neg[\Theta]sw_1, [\Theta]sw_2, [\Theta]sw_3, \neg[\Theta]relay, \neg[\Theta]light, \neg[\Theta]detect\} \cup \{[\Theta, N(\Theta)]toggle_1\}.$$

Consider $\Delta$ such that $\Delta_{AQ} = \mathcal{AB}_{AQ}\backslash\{[\Theta^{2+}]AQ_{detect}, [\Theta^{3+}]AQ_{detect}\}$ and $\Delta_{FA} = \mathcal{AB}_{FA}\backslash$ $\{[\Theta^+]FA_{\neg light}, [\Theta^+]FA_{\neg relay}, [\Theta^{2+}]FA_{sw_2}, [\Theta^{3+}]FA_{light}\}$. $\Delta$ is AD-plausible resulting in the next stable state being $\omega = \{sw_1, \neg sw_2, sw_3, relay, \neg light, \neg detect\}$ at $[\Theta^{4+}]$.

In addition, the set $\Delta' \subseteq \mathcal{AB}$, where $\Delta'_{AQ} = \mathcal{AB}_{AQ} \setminus \{[\Theta^{2+}]AQ_{detect}\}$ and $\Delta'_{FA} = \mathcal{AB}_{FA}\setminus \{[\Theta^+]FA_{\neg light}, [\Theta^+]FA_{\neg relay}, [\Theta^{2+}]FA_{sw_2}, [\Theta^{3+}]FA_{light}, [\Theta^{3+}]FA_{\neg detect}\}$, is also AD-plausible which results to the next stable state $\omega' = \{sw_1, \neg sw_2, sw_3, relay, \neg light, detect\}$ at $[\Theta^{4+}]$.

Moreover, the set $\Delta'' \subseteq \mathcal{AB}$, where $\Delta''_{AQ} = \mathcal{AB}_{AQ}$ and $\Delta''_{FA} = \mathcal{AB}_{FA}\setminus\{[N(\Theta)]FA_{\neg light}, [\Theta^+]FA_{\neg relay}, [\Theta^{2+}]FA_{sw_2}, [\Theta^{2+}]FA_{\neg detect}, [\Theta^{3+}]FA_{light}\}$, is also AD-plausible which also results to the next stable state $\omega'$ at $[\Theta^{4+}]$. In other words, the model implied by $\Delta'$ reflects a domain in which it takes the same amount of time for the *detect* to be on and the switch $sw_2$ to jump off. On the other hand, the model implied by $\Delta''$ reflects a domain in which the amount of time for the *detect* to be on is (approximately) equal to the amount of time for switch $sw_2$ to jump off and cause the *light* to be off as well. The *detect* implied by $\Delta$ is so insensitive that even though the *light* and the *relay* are on at the same time ($\Theta^{2+}$), the *relay* causes switch $sw_2$ to jump off and then leads to the *light* to be off as well but the *detect* is yet on.

**Example 2** (*continued.*) We re-formulate the action theory for this example in terms of our formalism:

$$\frac{[\tau]upL_1, [\tau]upL_2, [\tau]AQ_{open}}{[\tau^+]open \wedge \neg[\tau]FA_{\neg open}},$$

$$\frac{[\tau,v]flip_i, [\tau]upL_i}{\neg[v]upL_i \wedge \neg[\tau]FA_{upL_i}}(i=1,2),$$

$$\frac{[\tau,v]flip_i, \neg[\tau]upL_i}{[v]upL_i \wedge \neg[\tau]FA_{\neg upL_i}}(i=1,2),$$

$$\frac{[\tau,v]close, [\tau]upL_1, [\tau]upL_2}{\neg[v]open \wedge \neg[\tau]FA_{open}},$$

$$\frac{[\tau_1,\tau_2]hold\_closed, \neg[\tau_1]open, \tau_1 \le \tau \le \tau_2}{[\tau]held\_closed \wedge \neg[\tau]FA_{\neg held\_closed}},$$

$$\frac{[\tau]held\_closed}{\neg[\tau]AQ_{open}},$$

$$\frac{[\tau]\zeta, [\tau]FA_\zeta}{[\tau^+]\zeta}, \text{ where } \zeta \in \mathcal{F}^*.$$

The theory is described as follows:

$\Gamma = \{[\Theta]upL_1, [\Theta]upL_2, [\Theta]open, [\Theta]\neg held\_closed\} \cup \{[c_1, c_2]close, [c_2, c_3]hold\_closed, [c_4, c_5]flip_1\} \cup \{\Theta \le c_1 < c_2 \le c_4 < c_5 \le c_3\}$.

Consider $\Delta \subseteq \mathcal{AB}$ such that $\Delta_{AQ} = \mathcal{AB}_{AQ} \setminus \{[c]AQ_{open} \mid c_2 \le c \le c_3\}$ and $\Delta_{FA} = \mathcal{AB}_{FA} \setminus (\{[c_1]FA_{open}\} \cup \{[c]FA_{\neg held\_closed} \mid c_2 \le c \le c_3\} \cup \{[c_4]FA_{upL_1}\})$. $\Delta$ is AD-plausible and resulting in the following:

$[c_1]\{upL_1, upL_2, open, \neg held\_closed\},$
$[c_2]\{upL_1, upL_2, \neg open, held\_closed\},$
$[c_4]\{upL_1, upL_2, \neg open, held\_closed\},$
$[c_5]\{\neg upL_1, upL_2, \neg open, held\_closed\},$
$[c_3]\{\neg upL_1, upL_2, \neg open, held\_closed\}.$

**Example 3** (*continued.*) The domain description:

$$\frac{[\tau]sw_1, [\tau]sw_2, [\tau]AQ_{relay_1}}{[\tau^+]relay_1 \wedge \neg[\tau]FA_{\neg relay_1}},$$

$$\frac{[\tau]sw_1, \neg[\tau]sw_2, [\tau]AQ_{relay_2}}{[\tau^+]relay_2 \wedge \neg[\tau]FA_{\neg relay_2}},$$

$$\frac{[\tau]relay_1, [\tau]AQ_{\neg sw_2}}{\neg[\tau^+]sw_2 \wedge \neg[\tau]FA_{sw_2}},$$

$$\frac{[\tau]relay_2, [\tau]AQ_{sw_2}}{[\tau^+]sw_2 \wedge \neg[\tau]FA_{\neg sw_2}},$$

$$\frac{\neg[\tau]sw_i, [\tau,v]toggle_i}{[v]sw_i \wedge \neg[\tau]FA_{\neg sw_i}}(i=1,2),$$

$$\frac{[\tau]sw_i, [\tau,v]toggle_i}{\neg[v]sw_i \wedge \neg[\tau]FA_{sw_i}}(i=1,2),$$

$$\frac{[\tau]\zeta, [\tau]FA_\zeta}{[\tau^+]\zeta}, \text{ where } \zeta \in \mathcal{F}^*.$$

The state of the circuit in Figure 2 is captured by the following action theory:

$\Gamma = \{\neg[\Theta]sw_1, [\Theta]sw_2, \neg[\Theta]relay_1, \neg[\Theta]relay_2\} \cup \{[c_1, c_2]toggle_1, [c_3, c_4]toggle_1\} \cup \{\Theta \le c_1 < c_2 < c_3 < c_4\}$

Consider $\Delta \subseteq \mathcal{AB}$ such that $\Delta_{AQ} = \mathcal{AB}_{AQ}$ and $\Delta_{FA} = \mathcal{AB}_{FA} \setminus (\{[c_1]FA_{\neg sw_1}\} \cup \{[c_3]FA_{sw_1}\} \cup \{[c]FA_{\neg relay_1} \mid c_2 \le c \le c_3 \text{ and } c = (c_2)^{k+} \text{ where } k = 4i \text{ and } i = 0, 1, 2, \dots\} \cup \{[c]FA_{sw_2}, [c]FA_{relay_1} \mid c_2 \le c \le c_3 \text{ and } c = (c_2)^{k+} \text{ where } k = 4i+1 \text{ and } i = 0, 1, 2, \dots\} \cup \{[c]FA_{\neg relay_2} \mid c_2 \le c \le c_3 \text{ and } c = (c_2)^{k+} \text{ where } k = 4i+2 \text{ and } i = 0, 1, 2, \dots\} \cup \{[c]FA_{\neg sw_2}, [c]FA_{relay_2} \mid c_2 \le c \le c_3 \text{ and } c = (c_2)^{k+} \text{ where } k = 4i+3 \text{ and } i = 0, 1, 2, \dots\})$. $\Delta$ is AD-plausible resulting to several possible models for this domain depending on when switch $sw_1$ is toggled off:

$[c_1]\{\neg sw_1, sw_2, \neg relay_1, \neg relay_2\},$

$[c_2]\{sw_1, sw_2, \neg relay_1, \neg relay_2\}$,
$[c_3]\{sw_1, \pm sw_2, \mp relay_1, \pm relay_2\}$,
$[c_4]\{\neg sw_1, \pm sw_2, \mp relay_1, \pm relay_2\}$.

Remark:[20]

Here, it is also important to raise the question whether our solution to the ramification problem conflicts with the formulation of concurrent actions. When two actions $\alpha_1$ and $\alpha_2$ are concurrently performed: $\alpha_1$ causes an indirect effect $E$ and $\alpha_2$ triggers a non-terminating chains of effects. If the direct and indirect effects caused by $\alpha_2$ don't interfere with $\alpha_1$'s production of $E$ then the reasoner can reason about time points after the execution of $\alpha_1$ where $E$ holds for being an indirect effect of $\alpha_1$. These time points are of course unstable due to the non-terminating chains of indirect effects caused by $\alpha_2$. On the other hand, if the direct and indirect effects caused by $\alpha_2$ have the potential to prevent the former from producing $E$, the reasoning could be more complex. If it is observable that the direct effects of both actions take place at the same time then the concurrent occurrences of the two actions can be viewed as one single occurrence of one complex actions whose direct effects are a combination of the direct effects of the two actions. Further reasoning about ramifications will then be carried out as usual. On the other hand, when the reasoner herself is unsure about the temporal correlation between the direct effects of the two actions, every possible order of changes must be taken into consideration leading to a highly nondeterministic outcome about what fluent will hold in the future time points. However, such a situation can still be handled nicely by our formalisation. The only representation issue which is not addressed by our framework is about indirect effects with *duration*. As asserted above, given the state of the world at a time point $\tau$, we assume that all subsequent indirect effects will take place at the next time point $\tau^+$. This certainly fails to deal properly with indirect effects whose durations are *different*. A simple solution to this problem can be to associate one next time point operator for each possible indirect effect. This, however, will significantly complicate the representation.

## 7. Discussion and Future Work

We developed a uniform framework for reasoning about action using an argumentation-theoretic approach (more precisely, assumption-based approach). We have also presented how our framework copes with the frame, the qualification and the ramification problems in several sophisticated settings. We have shown how our framework can be naturally extended to become more expressive.

We also explored a new abstraction level which we believe to be an intermediate layer between the common sense knowledge and the scientific knowledge. Sophisticated domain knowledge as well as representation are, we argue, required to achieve an adequate underlying representation and reasoning process. Among the merits of this approach, we emphasise the following:

- *Non-monotonicity* is handled by assumptions and argumentation-theoretic approach.

---

20. The authors would like to acknowledge an anonymous referee for pointing out this very subtle issue.

- The *flexibility* of working with different kinds of information representation since there is no restriction on the syntax of the system.

- *Expressivity*: temporal information is explicitly represented. Thus the system is capable of capturing many important features of temporal reasoning.

As reviewed in the introductory section of this paper, numerous approaches to reasoning about action have been proposed. As such, much research related to these approaches and frameworks has evolved around the central problems addressed in this paper, namely the frame, qualification and the ramification problems. However, as non-monotonic reasoning-based formalisms for reasoning about actions were shown to be flawed by Hanks and McDermott (1987), many existing solutions of the frame problem are based on other approaches which are usually monotonic. Since one of the inherent properties of the argumentation-theoretic approach is monotonicity, few have attempted to solve the problems of reasoning about actions using this approach.

A framework for reasoning about actions under the argumentation-theoretic approach is independently proposed by Kakas, Miller, and Toni (1999, 2000, 2001). In their approach, the admissibility semantics is also employed to resolve conflicts between adversary arguments. To represent common sense knowledge, e.g. the common sense law of inertia, an order is imposed on the set of inference rules of the assumption-based framework. The computation of the arguments and their competing is performed on top of the so-called *proof trees*. Nodes on the proof trees are arguments which are sets of propositions.[21] Construction of proof trees is on the same level of hardness as known frameworks for argumentation-theoretic computation. While their framework allows the persistence of (inertial) fluents to be captured and dealt with, it is not very clear whether their formalisation can be extended to deal with other problems of reasoning about actions such as the qualification and the ramification problems.

More recently, Dimopoulos, Kakas, and Michael (2004), and Bracciali and Kakas (2004) extended Kakas et al.'s framework to deal with the ramification and the qualification problems. Essentially, in their solution to the ramification problem, the so-called *r-propositions* of the following form:[22]

$$L \text{ whenever } C$$

are added to the domain description. A resulting model for a given domain description will then be computed by first computing all possible indirect effects as the fixed point of repeated application of r-propositions, and then completing the model by allowing unaffected fluent literals to persist over time. It should be clear that the above representation of indirect effects is quite restrictive as it doesn't allow more complex expressions for the conditions of an indirect effects and the indirect effects themselves. These restriction is essentially due to their use of Answer Set Programming (ASP) as the underlying computation mechanism of their framework. On the other hand, their solution to the qualification problem lies in the use of default rules for representing the effects of actions. This approach is fairly similar to

---

21. In Kakas et al.'s (1999, 2000) terminologies, each node is a set of arguments which is equivalent to the notion of arguments in our exposition.

22. $L$ is a fluent literal and $C$ is a set of fluent literals which is essentially equivalent to a conjunction of fluent literals.

Thielscher's (2001) solution to the qualification problem discussed above. As their approach is essentially based on stability semantics, it's not clear how the argumentation-theoretic approach will bring about any benefit to their framework.

As discussed earlier in the paper, the key insight behind the solutions implemented in our framework is the exploitation of causality to drive the inference. It is causality that has helped throughout in our solutions to the frame and the qualification problems to provide the mechanisms to eliminate the anomalous models while retaining the intuitive models during the process of reasoning. This insight is certainly a more general version of the conclusions derived by Sandewall (1994) when investigating the reason behind the production of anomalous models in early approaches to the frame problem, namely the failure to distinguish between normal changes which are triggered by actions and abnomalous changes. Sandewall's insight has certainly originated the occlusion-based solution to the frame problem (Sandewall, 1994; Doherty, 1994; Gustafsson & Doherty, 1996). Roughly speaking, in this approach each action type is associated with a subset of fluents that are influenced by the action. Also, a predicate *Occlude* is introduced to allow this subset of fluents to be specified. When reasoning about dynamic domains, changes are minimised with the exception of the fluents specified by *Occlude*. In other words, *Occlude* distinguishes the normal changes (associated with the action types) from the anomalous changes and thus avoids unintended models from arising.

Regarding the qualification problem, Thielscher's (2001) solution to the problem of anomalous models in relation to the qualification problem shares the key insight of *causality* with our approach. However, Thielscher's framework is based on the standard semantics of Default Logic (with the initial theories are generated using circumscription). Thus, his approach does not deal with problem domains where standard semantics of Default Logic does not produce an extension. Furthermore, as discussed in the introductory section of our paper, in Thielscher's framework qualifications are taken over the executability conditions for the actions instead of over different effects of the actions.

In parallel to the causality-based insight to dynamic domains, other mechanisms for solving various problems of reasoning about action exist. For instance, a number of approaches employ the concept of *chronological ignorance* (Shoham, 1987, 1988) to tackle the problem of anomalous models. In general, causality-based frameworks and approaches based on chronological ignorance share the notion of directedness: when changes are minimised chronologically, causes are minimised instead of effects as causes are likely to precede effects. However, as a consequence, backward reasoning is blocked which prevents chronological ignorance-based approaches from dealing with surprises and expectation failures. Furthermore, non-deterministic actions or incomplete state knowledge are known to cause difficulty to chronological minimization. For a more detailed comparison between causality-based approaches and their counterparts that are based on chronological ignorance, the reader is referred to an article by Thielscher (2001). On the other hand, approaches that are based on *Motivated Action Theory* (MAT) (Amsterdam, 1991; Stein & Morgenstern, 1994) can be considered as a special case of the causality-based paradigm. MAT frameworks also advocate for the insight that an appropriate notion of causality is necessary when assuming away abnormalities. MAT frameworks, however, don't cope very well with the explanation problem and the ramification problem as pointed out by Jr. (1999) who also introduce an approach to improve MAT and overcome these problems.

Nonetheless, several issues still remain within our framework and will need further treatment to achieve an optimal solution. Firstly, this formalism may not be very suitable for the large-scale problems as too many assumptions will need to be taken into account. To address this problem, a localisation procedure is invented to guarantee that only an adequate sub-language will be used to capture the circumstance the agent is reasoning about. As a consequence, the set of assumptions will be restricted to those which are necessary to infer the conclusions the agent is interested. This idea is under development. Furthermore, while a uniform solution to all major problems of reasoning about action may be quite attractive especially regarding the toy scenarios such as those considered in this paper as well as in the literature, such a solution may not be pragmatic. By considering the locality account of reasoning, in particular of the assumptions used in default reasoning, promising solution can be achieved from both computational and representational points of view. We are undertaking further investigation towards this research direction.

A major limitation of our framework is the ability to deal with disjunctive axiomatisation of action occurrences. For instance, when the domain axiomatisation constains a disjunction about the action occurrences such as $[t_1, t_2]\alpha_1 \vee [t_1, t_2]\alpha_2$, then no plausible sets of assumptions, and accordingly, no CPMMs (resp. CPMQMs) will be produced to account for the effects of these actions. While an initial formalisation of our framework allowed complex expressions for action occurrences in the premises of inference rules which overcame this problem, the formalisation appeared to be too complex and some of the technical results could not be established. Further investigation, therefore, is needed to overcome this problem while avoiding to produce an overly complex formalisation.

Provided that our formalism has been designed to provide a general framework for reasoning about action, the following comes as a natural question: *To what extent the proposed approach can be used to formalise dynamic domains?* Sandewall (1994) suggests that one should systematise a framework for reasoning about action against some standard criteria to provide the formal indications about the expressiveness and capacity of a formalism. This is in contrast to the more traditional approach that had prevailed for many years in reasoning about action in which one tries to come up with some examples to show that no existing approaches are able to deal with such a scenario and claims that one's approach is better than others in which it can solve the proposed scenario.

Relative to Sandewall's standard criteria, our approach enjoys the following properties:

1. Our approach is non-inertia as it allows observations about the later state to correct the system's predictions about those states using some explanation mechanism through the use of assumptions.

2. Our formalism is able to deal with non-deterministic and concurrent events.

Another issue is the abduction problem which is also known in the literature as the explanation problem. For example, in the Stolen car scenario, by observing that the car disappeared from the parking lot where the reasoner had left it, an expectation failure arises. Most formalisms would try to accommodate this problem by introducing a stealing action as part of the vocabulary and try to bind the disappearance of the car to this action. This arguably is not very intuitive as there is no good reason why we should include this action in the vocabulary in the first place but not others such as the car being towed away by the police or a fairy turning the car into a pumpkin, etc. From a pragmatic point of view, some reasoners may just simply acquire more information (perhaps from the police) instead of confusing themselves with all kinds of explanations towards possible but uncertain causes. In other words, we effectively isolate the issues of deducing the new conclusions from the existing knowledge base from abducing the possible causes of some observations. This is of course closely related to Shanahan's approach in his IJCAI'95 paper (Shanahan, 1989) whose title clearly indicated that "*Prediction is Deduction but Explanation is Abduction*". We are also working on an assumption-based framework to solve the abduction problem.

### Acknowledgements

This work was performed when the first author was at the School of Computer Science and Engineering, University of New South Wales. The authors wish to thank other members of the Knowledge Systems Group, in paricular Dongmo Zhang and Maurice Pagnucco, and the anonymous reviewers of an earlier version of this paper for very helpful comments and suggestions that have significantly improve the quality as well as the readability of this paper. The first author was partially supported by an International Postgraduate Research Scholarship (IPRS) sponsored by the Australian government. The first author is presently supported by a by a DEST IAP grant (2004-2006, grant CG040014).

### Appendix A

**Theorem 1** *Let $D = \langle \mathcal{L}_D, \mathcal{R}, \mathcal{AB}, \Gamma \rangle$ be an S-domain. If $I$ is a CPMM of $D$ then $\Delta_{QF}^I$ is plausible.*

*Proof:* Suppose that $I$ is a CPMM of $D$,

(i) *we prove that $\Delta_{QF}^I$ is presumable:*

$\Delta_{QF}^I$ is $\mathcal{R}$-consistent since $I$ is a model of $D$. From Observation 2, we have $\Delta_{QF}^I$ is closed and does not attack itself. From Lemma 1, for each assumption $\delta \in AB$, if $\delta \notin \Delta_{QF}^I$ then $\delta$ is rejected by $\Delta_{QF}^I$.

(ii) *we prove that $Lr(\Delta_{QF}^I)$ is minimal:*

Suppose by way of contradiction that there exists a presumable set of assumptions $\Delta$ (wrt $D$) such that $Lr(\Delta) \subset Lr(\Delta_{QF}^I)$. Let $I_\Delta$ be the $\Delta$-relativised model of $D$. It is obvious that $I_\Delta$ is coherent. We will derive a contradiction by proving that $Occ^{I_\Delta} \subset Occ^I$:

(ii.1) $OA_D \subseteq Occ^I$: obvious as $OA_D = \{(\phi(\tau_1), \alpha, \phi(\tau_2)) \in \mathbb{T} \times \mathcal{A}_0 \times \mathbb{T} \mid \Gamma \models [\tau_1, \tau_2]\alpha\}$, and $I$ is a model of $D$.

(ii.2) $DAS(\Delta) \subseteq Occ^I$: Let $(t, da_{\neg l}, t^+) \in DAS(\Delta)$, then there do not exist any action $\alpha \in \mathcal{A}_0$ and $s \in \mathbb{T}$ such that

$$r = \frac{\Phi, [\tau_1, \tau_2]\alpha, [\tau_1, \tau_2]AQ_{\neg l}}{\neg[\tau_2]l \wedge \neg[\tau_1]FA_l} \in \mathcal{R}$$

and $I \models prem(r)[\tau_1/s, \tau_2/t^+]\}$. Thus, the assumption $\delta = [t]FA_l \in Lr(\Delta)$. Thus, $\delta \in Lr(\Delta_{QF}^I)$ (from the hypothesis.) From Lemma 2, $I \models [t, t^+]da_{\neg l}$. Thus $(t, da_{\neg l}, t^+) \in Occ^I$.

(ii.3) $Occ^I \nsubseteq Occ^{I_\Delta}$: Let $\delta = [t]FA_l \in Lr(\Delta_{QF}^I) \setminus Lr(\Delta)$ for some $t \in \mathbb{T}$ and $l \in \mathcal{F}^*$.

Since $I$ is a CPMM of $D$, from Lemma 2, $I \models [t, t^+]da_{\neg l}$. Thus $(t, da_{\neg l}, t^+) \in Occ^I$.

Moreover, $\delta \notin Lr(\Delta)$ **iff** either (a) $\delta \in \Delta$, or (b) there is $\alpha \in \mathcal{A}_0$ such that

$$r = \frac{\Phi, [\tau_1, \tau_2]\alpha, [\tau_1, \tau_2]AQ_{\neg l}}{\neg[\tau_2]l \wedge \neg[\tau_1]FA_l} \in \mathcal{R} \text{ and } \Gamma \cup \Delta \vdash_{\mathcal{R}} \Phi[\tau_1/t, \tau_2/t^+], [t, t^+]\alpha, [t, t^+]AQ_{\neg l}$$

**iff**, following the construction of $\Delta$-relativised models, $I_\Delta \not\models [t, t^+]da_{\neg l}$. Thus $(t, da_{\neg l}, t^+) \notin Occ^{I_\Delta}$. Therefore, $Occ^{I_\Delta} \subset Occ^I$.

Hence, $Lr(\Delta_{QF}^I)$ is minimal and $\Delta_{QF}^I$ is plausible. $\qquad\qquad \square$

**Theorem 2** *Let $D = \langle \mathcal{L}_D, \mathcal{R}, \mathcal{AB}, \Gamma \rangle$ be an S-domain and $\Delta \subseteq \mathcal{AB}$. $\Delta$ is plausible wrt $D$ iff $Mod_\Delta^S(D) \neq \emptyset$ and for each $I \in Mod_\Delta^S(D)$, $I$ is a CPMM of $D$.*

*Proof:*

($\Rightarrow$) Suppose that $\Delta \subseteq \mathcal{AB}$ is plausible wrt $D$. Then $\Gamma \cup \Delta$ is $\mathcal{R}$-consistent, i.e. $\Gamma \cup \Delta \not\vdash_{\mathcal{R}}$ **false**. From the construction of $\Delta$-relativised models, $Mod_\Delta^S(D) \neq \emptyset$.

For each $I = \langle h, \phi, \eta, \varepsilon_q, \varepsilon_f \rangle \in Mod_\Delta^S(D)$, we prove that $I$ is a CPMM of $D$.

(i) *$I$ is a coherent S-model of $D$:* obvious from the definition of $\Delta$-relativised models.

(ii) *$Occ^I$ is minimal:*

Assume the contrary, i.e. there is a non-empty set $\mathcal{M}_I = \{J \mid J$ is a coherent S-model of $D$ and $Occ^J \subset Occ^I\}$. Let $H \in \mathcal{M}_I$ such that there does not exist any model $J \in \mathcal{M}_I$ and $FA^H \subset FA^J$.

Consider the set of assumptions $\Delta_{QF}^H$:

(ii.a) $\Delta_{QF}^H$ *is presumable wrt $D$:*

- $\Delta_{QF}^H$ is closed and does not attack itself (since $H$ is a model of $D$);

- Let $\delta \in \mathcal{AB}$, if $\delta = [\tau]FA_l \notin \Delta_{QF}^H$ (for some $\tau \in \mathcal{TE}$ and $l \in \mathcal{F}^*$) and $\delta$ isn't rejected by $\Delta_{QF}^H$ then we can easily construct a model $J$ that interprets everything except $FA$ the same as $H$ and $FA^J = FA^H \cup \{(\phi(\tau), FA_l)\}$. Obviously, $J \in \mathcal{M}_I$ and $FA^J \subset FA^H$ which is a contradiction. Thus $\delta$ is rejected by $\Delta_{QF}^H$.

Therefore, $\Delta_{QF}^H$ is presumable.

(ii.b) $Lr(\Delta_{QF}^H) \subset Lr(\Delta)$:

$Occ^H \subset Occ^I$ since $H \in \mathcal{M}_I$.

Let $\delta \in Lr(\Delta_{QF}^H)$. Since $\mathcal{AB}_{AQ} \subseteq \Delta_{QF}^H$, $\delta = [t]FA_l \in \mathcal{AB}_{FA}$ for some $t \in \mathbb{T}$ and $l \in \mathcal{F}^*$. From the definition of leniently rejected assumptions, $\Gamma \cup \Delta_{QF}^H \not\vdash_{\mathcal{R}}$ **false** but $\Gamma \cup \Delta_{QF}^H \cup \{\delta\} \vdash_{\mathcal{R}}$ **false**. But then, from the definition of coherent models, $H \models [t, t^+]da_{\neg l}$ (since $H$ is a coherent model of $D$). Thus, $I \models [t, t^+]da_{\neg l}$.

From the construction of $\Delta$-relativised models, $[t]FA_l \notin \Delta$. Thus, $\delta$ is rejected by $\Delta$ (since $\Delta$ is plausible wrt $D$). Also from the construction of $\Delta$-relativised models, $\Gamma \cup \Delta \not\vdash_{\mathcal{R}} \neg\delta$. Therefore, $\delta \in Lr(\Delta)$, or $Lr(\Delta_{QF}^H) \subseteq Lr(\Delta)$. Now, $Occ^I = OA_D \cup DAS(\Delta)$. $OA_D \subseteq Occ^H$ since $H$ is a model of $D$. Thus, $Occ^I \setminus Occ^H = DAS(\Delta) \setminus Occ^H$. Let $(t, da_{\neg l}, t^+) \in DAS(\Delta) \setminus Occ^H$. From the construction of $\Delta$-relativised models, $\delta = [t]FA_l \in Lr(\Delta)$.

Suppose further that $\delta \in Lr(\Delta_{QF}^H)$. Then, $\delta \notin \Delta_{QF}^H$ (as $\Delta_{QF}^H$ is presumable). We construct a model $J$ in such a way that $J$ interprets everything except $FA$ the same as $H$.

From the definition of coherent models, $(t, da_{\neg l}, t^+) \notin Occ^H$ iff either (1) $H \not\models [t]l \wedge \neg[t^+]l$; or (2) there is $\alpha \in \mathcal{A}_0$, and

$$r = \frac{\Phi, [\tau_1, \tau_2]\alpha, [\tau_1, \tau_2]AQ_{\neg l}}{\neg[\tau_2]l \wedge \neg[\tau_1]FA_l} \in \mathcal{R}$$

and $\Gamma \cup \Delta \models prem(r)[\tau_1/t, \tau_2/t^+]$, i.e., $\Gamma \cup \Delta \models \Phi[\tau_1/t, \tau_2/t^+], [t, t^+]\alpha, [t, t^+]AQ_{\neg l}$. Since $\delta = [t]FA_l \in Lr(\Delta)$, condition (2) can not be satisfied.

Thus, $(t, da_{\neg l}, t^+) \notin Occ^H$ iff $H \not\models [t]l \wedge \neg[t^+]l$. In other words, it is consistent to add the assumption $[t]FA_l$ to the set of assumptions $\Delta_{QF}^H$. Or, from a model-theoretic point of view, to augment the denotation of $FA$ in $H$ with $(t, FA_l)$ and we can still obtain a coherent S-model $J$ such that $FA^H \subset FA^J$. But this is a contradiction. Hence, $\delta \notin Lr(\Delta_{QF}^H)$.

We have shown that $Lr(\Delta_{QF}^H) \subset Lr(\Delta)$.

From (ii.a) and (ii.b) we are led to the conclusion that $Occ^I$ is minimal. As a consequence, $I$ is a CPMM of $D$ (from (i) and (ii)).

($\Leftarrow$) Suppose that $Mod_\Delta^S(D) \neq \emptyset$ and for each $I \in Mod_\Delta^S(D)$, $I$ is a CPMM of $D$. We prove that $\Delta$ is plausible wrt $D$. Take an arbitrary model $I \in Mod_\Delta^S(D)$. Following Observation 4, $\Delta = \Delta_{QF}^I$. From Theorem 1 and the hypothesis that $I$ is a CPMM of $D$, we conclude that $\Delta$ is plausible wrt $D$. □

**Theorem 3** *Let $D = \langle \mathcal{L}_D, \mathcal{R}, \mathcal{AB}, \Gamma \rangle$ be an S-domain. Furthermore, suppose that $CPMM(D)$ is the set of CPMMs of $D$ and $Plaus(D)$ is the set of plausible sets of assumptions of $D$, then $CPMM(D) = \bigcup_{\Delta \in Plaus(D)} Mod_\Delta^S(D)$.*

*Proof:*

($\supseteq$) Let $\Delta \in Plaus(D)$, then $Mod_\Delta^S(D) \subseteq CPMM(D)$ (Following Theorem 2). Therefore, $\bigcup_{\Delta \in Plaus(D)} Mod_\Delta^S(D) \subseteq CPMM(D)$.

($\subseteq$) Let $I \in CPMM(D)$, then $\Delta_{QF}^I \in Plaus(D)$ (Following Theorem 1).

We prove that $I \in Mod_{\Delta_{QF}^I}(D)$ based on Definition 5.4:

1) $I$ is an S-model of $D$,

2) for each $\delta = [\tau]FA_l \in \mathcal{AB}_{FA}$ (for some $\tau \in \mathcal{TE}$ and $l \in \mathcal{F}^*$), $\delta \in \Delta_{QF}^I$ iff $(\phi(\tau), FA_l) \in FA^I$ (Following the definition of $\Delta_{QF}^I$ - Definition 5.3)

3) We prove that $Occ^I = OA_D \cup DAS(\Delta_{QF}^I)$:

(3.$\supseteq$)

- $OA_D = \{(\phi(\tau_1), \alpha, \phi(\tau_2)) \in \mathbb{T} \times \mathcal{A}_0 \times \mathbb{T} \mid \Gamma \models [\tau_1, \tau_2]\alpha\} \subseteq Occ^I$ (as $I$ is a model of $D$),

- $(t, da_{\neg l}, t^+) \in DAS(\Delta_{QF}^I)$.

  From Definition 5.4,

  (i) $\delta = [t]FA_l \notin \Delta_{QF}^I$, and

  (ii) there exists no action $\alpha \in \mathcal{A}_0$ such that:

$$r = \frac{\Phi, [\tau_1, \tau_2]\alpha, [\tau_1, \tau_2]AQ_{\neg l}}{\neg[\tau_2]l \wedge \neg[\tau_1]FA_l} \in \mathcal{R}$$

and $\Gamma \cup \Delta \models \Phi[\tau_1/t, \tau_2/t^+], [t, t^+]\alpha, [t, t^+]AQ_{\neg l}$.

From Lemma 1, $\delta \notin \Delta^I_{QF}$ iff $\delta$ is rejected by $\Delta^I_{QF}$ iff — as $I$ is a coherent model — either (a) there are $\alpha \in \mathcal{A}_0$ and $s \in \mathbb{T}$ such that

$$r = \frac{\Phi, [\tau_1, \tau_2]\alpha, [\tau_1, \tau_2]AQ_{\neg l}}{\neg[\tau_2]l \wedge \neg[\tau_1]FA_l} \in \mathcal{R}$$

and $I \models prem(r)[\tau_1/s, \tau_2/t^+]$,

or (b) $I \models [t, t^+]da_{\neg l}$.

Since (a) violates condition (ii) above, (b) must be the case. Thus $(t, da_{\neg l}, t^+) \in Occ^I$.

$\Rightarrow DAS(\Delta^I_{QF}) \subseteq Occ^I$.

$\Rightarrow Occ^I \supseteq OA_D \cup DAS(\Delta^I_{QF})$.

(3.$\subseteq$) Suppose that $Occ^I \not\subseteq OA_D \cup DAS(\Delta^I_{QF})$. From (3.$\supseteq$), we have $Occ^I \supset OA_D \cup DAS(\Delta^I_{QF})$. Since $\Delta^I_{QF}$ is plausible, there exists a model $J \in Mod^S_{\Delta^I_{QF}}(D)$ such that $J$ is a CPMM of $D$. Following Definition 5.4, $Occ^J = OA_D \cup DAS(\Delta^I_{QF})$ which is a proper subset of $Occ^I$. Contradiction! Therefore, $Occ^I \subseteq OA_D \cup DAS(\Delta^I_{QF})$.

Thus we have shown that $I \in Mod^S_{\Delta^I_{QF}}(D)$,

$\Rightarrow M \in \bigcup_{\Delta \in Plaus(D)} Mod^S_\Delta(D)$ (Since $\Delta^I_{QF} \in Plaus(D)$)

$\Rightarrow CPMM(D) \subseteq \bigcup_{\Delta \in Plaus(D)} Mod^S_\Delta(D)$.

Therefore, $CPMM(D) = \bigcup_{\Delta \in Plaus(D)} Mod^S_\Delta(D)$. □

**Theorem 4** *Let $D$ be a Q-domain. If $I$ is a CPMQM of $D$ then $\Delta^I_{QF}$ is Q-plausible wrt $D$.*

*Proof:*

Suppose that $I$ is a CPMQM of $D$,

(i) it's easy to verify that $\Delta^I_{QF}$ is semi–Q-plausible wrt $D$: similar to the proof of theorem 1, but using Lemma 3 instead of Lemma 1 and Lemma 2.

(ii) $\Delta^I_{AQ}$ is maximal:

Assume the contrary, i.e. there exists a set of assumptions $\Delta$ such that $\Delta$ is semi–Q-plausible wrt $D$ and $\Delta^I_{AQ} \subset \Delta_{AQ}$. Since $\Delta$ is presumable, $\Gamma \cup \Delta$ is $\mathcal{R}$-consistent. Let $I_\Delta$ be the $\Delta$-relativised model of $D$. Obviously, $I_\Delta$ is a coherent Q-model of $D$. It's easy to verify that $Occ^{I_\Delta}$ is minimal because otherwise a coherent Q-model $J$ (of $D$) can be constructed such that $Occ^J \subset Occ^{I_\Delta}$. We have $Occ^{I_\Delta} = OA_D \cup DAS(\Delta)$ and $OA_D \subseteq Occ^J$ since $J$ is a model of $D$. Thus there exists $(t, da_{\neg\varphi}, t^+) \in DAS(\Delta) \setminus Occ^J$. But then $\Delta^J_{QF}$ is presumable wrt $D$ and $Lr(\Delta^J_{QF}) \subset Lr(\Delta)$ which is a contradiction. Thus, $Occ^{I_\Delta}$ is minimal. But then the model $I_\Delta$ is a PMQM of $D$ and $AQ^{I_\Delta} \subset AQ^I$ which contradicts with the fact that $I$ is a CPMQM of $D$. Therefore, $\Delta^I_{AQ}$ is maximal.

(iii) $\Delta^I_{FA}$ is maximal (relative to (i) and (ii)):

Assume the contrary, i.e. there exists a set of assumptions $\Delta$ such that $\Delta$ is semi–Q-plausible wrt $D$ and $\Delta_{AQ}$ is maximal and $\Delta^I_{FA} \subset \Delta_{FA}$. Let $I_\Delta$ be the $\Delta$-relativised model of $D$. Similar to the proof in part (ii), we can easily verify that $I_\Delta$ is a PMQM of $D$. As the fact that $\Delta_{AQ}$ is maximal and $\Delta_{FA}^I \subset \Delta_{FA}$ contradicts with the given hypothesis that $I$ is a CPMQM of $D$, we conclude that $\Delta_{FA}^I$ is maximal. Therefore, $\Delta_{QF}^I$ is Q-plausible wrt $D$. $\square$

**Theorem 5** *Let $D = \langle \mathcal{L}_D, \mathcal{R}, \mathcal{AB}, \Gamma \rangle$ be a Q-domain and $\Delta \subseteq \mathcal{AB}$. $\Delta$ is Q-plausible wrt $D$ iff $Mod_\Delta^Q(D) \neq \emptyset$ and for each $I \in Mod_\Delta^Q(D)$, $I$ is a CPMQM of $D$.*

*Proof:*

($\Rightarrow$) Suppose that $\Delta$ is Q-plausible wrt $D$.

As $\Delta$ is presumable, $\Gamma \cup \Delta$ is $\mathcal{R}$-consistent. Following the construction of $\Delta$-relativised models, $Mod_\Delta^Q(D) \neq \emptyset$.

Let $I \in Mod_\Delta^Q(D)$:

(i) it's easy to verify that $I$ is coherent from the construction of $\Delta$-relativised models.

(ii) $Occ^I$ is minimal:

Assume the contrary, i.e. the set $\Sigma_I = \{\sigma \mid \sigma$ is a coherent Q-model of $D$ and $Occ^\sigma \subset Occ^I\}$ is non-empty.

Let $J \in \Sigma_I$ such that the following conditions are satisfied:

1. there does not exist any model $\sigma \in \Sigma_I$ such that $Occ^\sigma \subset Occ^J$.

2. $AQ^J$ is maximal relative to 1.

3. $FA^J$ is maximal relative to 1. and 2.

Consider the set of assumptions $\Delta_{QF}^J$: Obviously, $\Delta_{QF}^J$ is closed and does not attach itself. For each $\delta \in \mathcal{AB}$, if $\delta \notin \Delta_{QF}^J$ then $\delta$ is rejected by $\Delta_{QF}^J$, otherwise it would violate the maximality of $\Delta_{AQ}^J$ and $\Delta_{FA}^J$. Thus, $\Delta_{QF}^J$ is presumable.

Remark that $Occ^J \subset Occ^I$. Besides, $Occ^I = OA_D \cup DAS(\Delta)$. But $OA_D \subseteq Occ^J$ as $J$ is a model of $D$. Thus there exists $(t, da_{\neg\varphi}, t^+) \in DAS(\Delta) \setminus Occ^J$. But then $\Delta_{QF}^J$ is presumable wrt $D$ and $Lr(\Delta_{QF}^J) \subset Lr(\Delta)$ which is a contradiction. Thus, $Occ^I$ is minimal. As a consequence, $I$ is a PMQM of $D$.

(iii) $AQ^I$ is maximal (relative to (i) and (ii)):

Assume the contrary, i.e. the set $\Sigma_I = \{\sigma \mid \sigma$ is a PMQM of $D$ and $AQ^I \subset AQ^\sigma\}$ is non-empty.

Let $J \in \Sigma_I$ such that the following conditions are satisfied:

1. $FA^J$ is maximal; and

2. $AQ^J$ is maximal relative to 1.

We can prove that that $\Delta_{QF}^J$ is semi-Q-plausible (i.e. presumable wrt $D$ and $Lr_{FA}(\Delta_{QF}^J)$ is minimal):

That $\Delta_{QF}^J$ is presumable wrt $D$ is easy to verify.

It's also easy to verify that $Lr_{FA}(\Delta_{QF}^J)$ is minimal as $J$ is coherent and $Occ^J$ is minimal. This would guarantee that the set of occurrences of dummy actions in $J$ is minimised and as a consequence the set of leniently rejected frame assumptions is also minimised. Formally, if a presumable set of assumptions $\Pi$ is such that $Lr_{FA}(\Pi) \subset Lr_{FA}(\Delta_{QF}^J)$ then the $\Delta$-relativised model $I_\Pi$ is a coherent Q-model and $Occ^{I_\Pi} \subset Occ^J$, which is a contradiction with the fact that $J \in \Sigma_I$ and thus $J$ is a PMQM of $D$.

Thus, $\Delta_{QF}^J$ is semi-Q-plausible wrt $D$. But, $\Delta_{AQ} \subset \Delta_{AQ}^J$ which is a contradiction.

Therefore, we have shown that $AQ^I$ is maximal (relative to (i) and (ii)).

(iv) Now, we can prove that $FA^I$ is maximal (relative to (i), (ii) and (iii)):

Assume the contrary, i.e. there exists a PMQM $J$ of $D$ such that $FA^I \subset FA^J$.

Among those models satisfying the above condition, we choose a model $K$ such that there does not exist any PMQM $K'$ of $D$ such that $FA^K \subset FA^{K'}$. Thus, $K$ is a CPMQM of $D$.

Following Theorem 4, we have the set of assumptions $\Delta_{QF}^K$ is Q-plausible and $\Delta_{FA} \subset \Delta_{FA}^K$ since $FA^I \subset FA^K$. Contradiction with the hypothesis that $\Delta$ is Q-plausible.

From (i), (ii), (iii) and (iv) we are led to the conclusion that $I$ is a canonical prioritised minimal Q-model of $D$.

($\Leftarrow$) Suppose that $Mod_\Delta^Q(D) \neq \emptyset$ and $I$ is a canonical prioritised minimal model of $D$ for each $I \in Mod_\Delta^Q(D)$, we prove that $\Delta$ is plausible.

Take an arbitrary model $I \in Mod_\Delta^Q(D)$. Following Observation 5, $\Delta = \Delta_{QF}^I$. From Theorem 4 and the hypothesis that $I$ is a canonical prioritised minimal Q-model of $D$, we conclude that $\Delta$ is Q-plausible. $\qquad\square$

**Theorem 6** *Let $D$ be a Q-domain. Furthermore, suppose that $CPMQM(D)$ is the set of CPMQMs of $D$ and $Plaus^Q(D)$ is the set of Q-plausible sets of assumptions of $D$, then $CPMQM(D) = \bigcup_{\Delta \in Plaus^Q(D)} Mod_\Delta^Q(D)$.*

*Proof:* Similar to the proof of Theorem 3. $\qquad\square$

**Theorem 7** *Let $\sigma = \langle \mathcal{T}, \mathcal{F}, \mathcal{A} \rangle$ be a signature and $D_0 = \langle \mathcal{L}_D, \mathcal{R}, \mathcal{AB}, \emptyset \rangle$ a SSD and $\alpha \in \mathcal{A}_0$. Suppose that $w \in IS$. Define a domain description $D = \langle \mathcal{L}_D, \mathcal{R}, \mathcal{AB}, \Gamma \rangle$, where $\Gamma = \{[\Theta]\varphi \mid \varphi \in w\} \cup \{[\Theta, \Theta^+]\alpha\}$. A set $\Delta \subseteq \mathcal{AB}$ of assumptions is AD-plausible wrt $D$ iff for each model $M$ of $E_D(\Delta)$, $[M]^{N_M(\Theta)} \in Trans_D^\alpha([M]^\Theta)$, i.e. $[M]^{N_M(\Theta)}$ belongs to the state transition from $[M]^\Theta$ in $D$ according to $\alpha$.*

*Proof:*

($\Rightarrow$) Suppose that $\Delta \subseteq \mathcal{AB}$ is AD-plausible wrt $D$, we prove that for each model $M$ of $E_D(\Delta)$, $[M]^{N_M(\Theta)} \in Trans_D^\alpha([M]^\Theta)$.

Let $M \in Mod(E_D(\Delta))$, as $M \models \Gamma$, we have $M \models [\Theta]\varphi$ iff $\varphi \in w$. Thus $w = [M]^\Theta$.

We prove that $[M]^{N_M(\Theta)} \in Trans_D^\alpha(w)$, i.e. there exists a sequence $\omega_1, \ldots, \omega_n \in IS$ such that $(w, \alpha, \omega_1) \in Res_D$ and $[M]^{N_M(\Theta)} = \omega_n$ and $(\omega_i, \omega_{i+1}) \in Causes_D$ for each $1 \leq i < n$.

If $\Upsilon_\alpha^w = \emptyset$: to the reasoner's knowledge, $\alpha$ is not applicable in the instantwise state $w$ due to either non-executability of $\alpha$ in $w$, or $\alpha$ does not bring about any effects concerned to the reasoner. Thus, $[M]^{\Theta^+} = w$. Of course, $(w, \alpha, [M]^{\Theta^+}) \in Res_D$.

If $\Upsilon_\alpha^w \neq \emptyset$: let $\Omega = \{r \in \mathcal{R}_A \mid M \models prem(r) \text{ and } M \models cons(r)\}$, we prove that $\Omega$ is a possible application of $\alpha$ in $w$.

Apparently, $\Omega \subseteq \Upsilon_\alpha^w$. $\Omega$ is maximal since otherwise we can construct a model $M'$ such that $M'$ satisfies the qualification assumption of the additional action description rule. But it means $M'$ is a PMQM model of $D$ such that $AQ^M \subset AQ^{M'}$. Thus $M$ is not a CPMQM of $D$. Following Theorem 5, $\Delta$ is not Q-plausible wrt $D$. Contradiction.

We now prove that there does not exists any instantwise state $S$ such that $CONS_A(\Omega) \subseteq S$ and $[M]^{\Theta^+} \setminus w \subset S \setminus w$.

Assume the contrary, i.e. there exists a fluent literal $\varphi \in \mathcal{F}^*$ such that $\varphi \in (S \setminus w) \setminus ([M]^{\Theta^+} \setminus w)$.

Then $M \not\models [\Theta]FA_\varphi$ since $M$ is a model of $D$.

Construct a model $M'$ in such a way that $M'$ interprets everything the same as $M$ except $FA$, and $FA^{M'} = FA^M \cup \{(\Theta, FA_\varphi)\}$. Obviously, $M'$ is a PMQM of $D$ and $AQ^{M'} = AQ^M$, but $FA^M \subset FA^{M'}$. Thus $M$ is not a CPMQM of $D$. Following Theorem 5, $\Delta$ is not Q-plausible wrt $D$. Contradiction. Therefore, $(w, \alpha, [M]^{\Theta^+}) \in Res_D$.

($\Leftarrow$) Suppose that for each model $M$ of $E_D(\Delta)$, $[M]^{N_M(\Theta)} \in Trans_D^\alpha([M]^\Theta)$, we prove that $\Delta \subseteq \mathcal{AB}$ is AD-plausible wrt $D$ which is obvious. □

# References

Amsterdam, J. B. (1991). Temporal reasoning and narrative conventions. In Allen, J. F., Fikes, R., & Sandewall, E. (Eds.), *KR'91: Principles of Knowledge Representation and Reasoning*, pp. 15–21, Cambridge, MA. Morgan Kaufmann.

Baker, A. B. (1989). A simple solution to the Yale Shooting problem. In Brachman, R. J., Levesque, H. J., & Reiter, R. (Eds.), *KR'89: Principles of Knowledge Representation and Reasoning*, pp. 11–20, San Mateo, California. Morgan Kaufmann.

Baral, C. (1995). Reasoning about actions: Non-deterministic effects, constraints, and qualification. In *International Joint Conference on Artificial Intelligence*.

Bondarenko, A., Dung, P. M., Kowalski, R. A., & Toni, F. (1997). An abstract, argumentation-theoretic approach to default reasoning. *Artificial Intelligence Journal*, *93*, 63–101.

Bracciali, A., & Kakas, A. C. (2004). Frame consistency: computing with causal explanations. In *10th International Workshop on Non-Monotonic Reasoning (NMR 2004)*, pp. 79–87.

Castilho, M. A., Gasquet, O., & Herzig, A. (1999). Formalizing action and change in modal logic I: The frame problem. *Journal of Logic and Computation*, *9(5)*, 701–735.

Dimopoulos, Y., Kakas, A. C., & Michael, L. (2004). Reasoning about actions and change in answer set programming. In *International Conference on Logic Programming and Nonmonotonic Reasoning - LPNMR' 04*, pp. 61–73.

Doherty, P. (1994). Reasoning about action and change using occlusion. In *European Conference on Artificial Intelligence*, pp. 401–405.

Doherty, P., & Kvarnström, J. (1998). Tackling the qualification problem using fluent dependency constraints: Preliminary report. In *5th Workshop on Temporal Representation and Reasoning - TIME*, pp. 97–104.

Doyle, J. (1979). A truth maintenance system. *Artificial Intelligence Journal*, *12(3)*, 231–272.

Drakengren, T., & Bjäreland, M. (1999). Reasoning about action in polynomial time. *Artificial Intelligence Journal*, *115*, 1–24.

Fikes, R., & Nilsson, N. J. (1971). STRIPS: A new approach to the application of theorem proving to problem solving. *Artificial Intelligence Journal*, *2(3/4)*, 189–208.

Foo, N. Y., Zhang, D., Vo, Q. B., & Peppas, P. (2001). Circumscriptive models and automata. In Thielscher, M., & Williams, M.-A. (Eds.), *Workshop on Non-monotonic Reasoning, Action and Change - colocated with IJCAI-01*.

Gelfond, M., & Lifschitz, V. (1998). Action languages. *Electronic Transactions on AI*, *3(16)*, 193–210.

Ginsberg, M. L., & Smith, D. E. (1988). Reasoning about action I: A possible worlds approach. *Artificial Intelligence Journal*, *35(2)*, 165–196.

Giunchiglia, E., Kartha, G. N., & Lifschitz, V. (1997). Representing action: Indeterminacy and ramifications. *Artificial Intelligence Journal*, *95(2)*, 409–438.

Giunchiglia, E., & Lifschitz, V. (1998). An action language based on causal explanation: Preliminary report. In *National Conference on Artificial Intelligence*, pp. 623–630.

Green, C. (1969). Application of theorem proving to problem solving. In *International Joint Conference on Artificial Intelligence*, pp. 219–240.

Gustafsson, J., & Doherty, P. (1996). Embracing occlusion in specifying the indirect effects of actions. In *Principles of Knowledge Representation and Reasoning*, pp. 87–98.

Haas, A. R. (1987). The case for domain-specific frame axioms. In *The frame problem in artificial intelligence: proc. of 1987 workshop*. Morgan Kaufmann.

Hanks, S., & McDermott, D. (1987). Nonmonotonic logic and temporal projection. *Artificial Intelligence Journal*, *33(3)*, 379–412.

Jr., C. L. O. (1999). Explanatory update theory: Applications of counterfactual reasoning to causation. *Artificial Intelligence Journal*, *108(1-2)*, 125–178.

Kakas, A. C., Miller, R., & Toni, F. (1999). An argumentation framework of reasoning about actions and change. In *International Conference on Logic Programming and Nonmonotonic Reasoning - LPNMR' 99*, pp. 78–91.

Kakas, A. C., Miller, R., & Toni, F. (2000). E-res - a system for reasoning about actions, events and observations. In Baral, C., & Truszczynski, M. (Eds.), *International Workshop on Non-Monotonic Reasoning, Special Session on System Descriptions and Demonstration - NMR'2000*.

Kakas, A. C., Miller, R., & Toni, F. (2001). E-res: Reasoning about actions, events and observations. In *International Conference on Logic Programming and Nonmonotonic Reasoning - LPNMR' 01*, pp. 254–266.

Kautz, H. (1986). The logic of persistence. In *National Conference on Artificial Intelligence*.

Kowalski, R., & Sergot, M. J. (1986). A logic-based calculus of events. *New Generation Computing*, *4*, 67–95.

Kowalski, R. (1992). Database updates in the event calculus. *Journal of Logic Programming*, *12*, 121–146.

Kushmerick, N. (1996). Cognitivism and situated action: two views on intelligent agency. *Computers and Artificial Intelligence*, *15(5)*.

Lifschitz, V. (1987). On the semantics of STRIPS. In Georgeff, & Lansky (Eds.), *Reasoning about Actions and Plans*. Morgan Kauffman, Los Altos.

Lin, F. (1995). Embracing causality in specifying the indirect effects of actions. In *International Joint Conference on Artificial Intelligence*.

Lin, F. (1996). Embracing causality in specifying the indeterminate effects of actions. In *National Conference on Artificial Intelligence*, pp. 670–676.

Lin, F., & Reiter, R. (1994). State constraints revisited. *Journal of Logic and Computation, 4(5)*, 655–678.

Lin, F., & Shoham, Y. (1995). Provably correct theories of action. *Journal of the ACM, 42(2)*, 293–320.

McCain, N., & Turner, H. (1995). A causal theory of ramifications and qualifications. In *International Joint Conference on Artificial Intelligence*.

McCain, N., & Turner, H. (1997). Causal theories of action and change. In *National Conference on Artificial Intelligence*, pp. 460–465.

McCarthy, J. (1977). Epistemological problems of artificial intelligence. In *International Joint Conference on Artificial Intelligence*, pp. 555–562.

McCarthy, J. (1980). Circumscription - a form of non-monotonic reasoning. *Artificial Intelligence Journal, 13(1-2)*, 27–39.

McCarthy, J. (1986). Applications of circumscription to formalizing common sense knowledge. *Artificial Intelligence Journal, 26(3)*, 89–116.

McCarthy, J., & Hayes, P. (1969). Some philosophical problems from the standpoint of artificial intelligence. In Michie, D., & Meltzer, B. (Eds.), *Machine Intelligence 4*. Edinburgh University Press.

McDermott, D. V. (1987). We've been framed: Or why ai is innocent of the frame problem. In Pylyshyn, Z. (Ed.), *The Robots Dilemma: The Frame Problem in Artificial Intelligence*, pp. 113–122. Ablex.

McDermott, D. V., & Doyle, J. (1980). Non-monotonic logic I. *Artificial Intelligence Journal, 13(1-2)*, 41–72.

Moore, R. C. (1985). Semantical considerations on nonmonotonic logic. *Artificial Intelligence Journal, 25(1)*, 75–94.

Morris, P. H. (1988). The anomalous extension problem in default reasoning. *Artificial Intelligence Journal, 35(3)*, 383–399.

Pednault, E. (1989). ADL: Exploring the middle ground between STRIPS and the situation calculus. In *KR'89: First International Conference on Principles of Knowledge Representation and Reasoning*, pp. 324–332. Morgan Kaufmann.

Poole, D. (1988). A logical framework for default reasoning. *Artificial Intelligence Journal, 36(1)*, 27–47.

Reiter, R. (1980). A logic for default reasoning. *Artificial Intelligence Journal, 13*, 81–132.

Reiter, R. (1991). The frame problem in the situation calculus: A simple solution (sometimes) and a completeness result for goal regression. In Lifschitz, V. (Ed.), *AI and Mathematical Theory of Computation: Papers in Honor of John McCarthy*, pp. 418–420. Academic Press.

Sandewall, E. (1994). *Features and Fluents*. Oxford University Press, Oxford.

Schubert, L. (1990). Monotonic solution of the frame problem in the situation calculus; an efficient method for worlds with fully specified actions. In Kyburg, H., Loui, R., & Carlson, G. (Eds.), *Knowledge Representation and Defeasible Reasoning*, pp. 23–67. Kluwer Academic Publishers, Dordrecht.

Shanahan, M. (1989). Prediction is deduction but explanation is abduction. In *International Joint Conference on Artificial Intelligence*, pp. 1055–1060.

Shanahan, M. (1997). *Solving the Frame Problem: A Mathematical Investigation of the Common Sense Law of Inertia*. MIT Press, Cambridge, Massachussets.

Shanahan, M. (1999). The ramification problem in the event calculus. In *International Joint Conference on Artificial Intelligence*, pp. 140–146.

Shoham, Y. (1987). *Reasoning about Change*. MIT Press, Cambridge, MA.

Shoham, Y. (1988). Chronological ignorance: Experiments in nonmonotonic temporal reasoning. *Artificial Intelligence Journal*, *36*, 279–331.

Stein, L. A., & Morgenstern, L. (1994). Motivated action theory: a formal theory of causal reasoning. *Artificial Intelligence Journal*, *71(1)*, 1–42.

Thielscher, M. (1997). Ramification and causality. *Artificial Intelligence Journal*, *89*, 317–364.

Thielscher, M. (1999). From situation calculus to fluent calculus: State update axioms as a solution to the inferential frame problem. *Artificial Intelligence Journal*, *111(1-2)*, 277–299.

Thielscher, M. (2001). The qualification problem: A solution to the problem of anomalous models. *Artificial Intelligence Journal*, *131(1-2)*, 1–37.

Turner, H. (1997). Representing actions in logic programs and default theories: A situation calculus approach. *Journal of Logic Programming*, *31(1-3)*, 245–298.

Vo, Q. B., & Foo, N. Y. (2001). Solving the qualification problem. In *Australian Joint Conference on Artificial Intelligence*, pp. 519–531.

Vo, Q. B., & Foo, N. Y. (2002). Solving the ramification problem: Causal propagation in an argumentation-theoretic approach. In *7th Pacific Rim International Conference on Artificial Intelligence - PRICAI2002*, pp. 49–59.

Zhang, D., & Foo, N. Y. (2002). Interpolation properties of action logic: Lazy-formalization to the frame problem. In Flesca, S., Greco, S., Leone, N., & Ianni, G. (Eds.), *Logics in Artificial Intelligence, European Conference, JELIA 2002*, pp. 357–368.